\documentclass[journal]{IEEEtran}
\usepackage{cite}

\ifCLASSINFOpdf
\else
\fi
\usepackage[cmex10]{amsmath}
\usepackage{algorithm}
\usepackage{algorithmic}

\usepackage{array}
\usepackage[caption=false,font=footnotesize]{subfig}
\usepackage{url}

\usepackage{graphicx,listings,psfrag,amsfonts,cases}

\usepackage{pst-sigsys}
\usepackage{pstricks,pst-node,pst-tree}

\usepackage{amsmath,amssymb,amsfonts,graphicx,nicefrac,mathtools,bbm}
\usepackage{amsthm}

\newcommand{\BIT}{\begin{itemize}}
\newcommand{\EIT}{\end{itemize}}
\newcommand{\BNUM}{\begin{enumerate}}
\newcommand{\ENUM}{\end{enumerate}}
 
\newcommand\mbb[1]{\mathbb{#1}}

\def\reals{\mathbb{R}} 
\def\complex{\mathbb{C}} 
\renewcommand{\exp}[1]{\operatorname{exp}\left(#1\right)} 
\def\indic#1{\mbb{I}\left({#1}\right)} 

\def\E{\mathbb{E}} 

\def\P{\mathbb{P}} 

\def\Unif{\textnormal{Unif}}

\newtheorem{theorem}{Theorem}
\newtheorem{lemma}{Lemma}
\newtheorem{assumption}{Assumption}
\newtheorem*{remark}{Remark}

\providecommand{\re}{\mathop\mathrm{Re}}
\providecommand{\im}{\mathop\mathrm{Im}}

\usepackage{multirow}

\newcommand{\putFig}[3]{
        \begin{figure}[ht!] 
 		\centering
 		\includegraphics[width=#3]{#1}
		  \caption{#2}
                \label{fig:#1}
        \end{figure}}

\def\DI{\Delta I}
\def\DV{\Delta V}

\def\Dv{\Delta v}

\newcommand{\RN}[1]{%
  \textup{\uppercase\expandafter{\romannumeral#1}}%
}

\usepackage{color}
\newcommand{\rev}[1]{#1}
\newcommand{\rmtxt}[1]{}

\begin{document}
%
\title{Urban MV and LV Distribution Grid Topology Estimation via Group Lasso}

\author{Yizheng~Liao,~\IEEEmembership{Student Member,~IEEE,}
        Yang~Weng,~\IEEEmembership{Member,~IEEE,}
        Guangyi~Liu,~\IEEEmembership{Senior Member,~IEEE,}
        Ram~Rajagopal,~\IEEEmembership{Member,~IEEE}
\thanks{Y. Liao and R. Rajagopal are with Department of Civil and Environmental Engineering, Stanford University, Stanford, CA, 94305 USA e-mail: \{yzliao,ramr\}@stanford.edu. Y. Weng is with School of Electrical, Computing, and Energy Engineering, Arizona State University, Tempe, AZ, 85287 USA e-mail: yang.weng@asu.edu. G. Liu is with State Grid GEIRI North America, Santa Clara, CA, 95054 USA e-mail: guangyiliu2010@gmail.com.}}


\maketitle

\begin{abstract}
The increasing penetration of distributed energy resources poses numerous reliability issues to the urban distribution grid. The topology estimation is a critical step to ensure the robustness of distribution grid operation. However, the bus connectivity and grid topology estimation are usually hard in distribution grids. For example, it is technically challenging and costly to monitor the bus connectivity in urban grids, e.g., underground lines. It is also inappropriate to use the radial topology assumption exclusively because the grids of metropolitan cities and regions with dense loads could be with many mesh structures. To resolve these drawbacks, we propose a data-driven topology estimation method for MV and LV distribution grids by only utilizing the historical smart meter measurements. Particularly, a probabilistic graphical model is utilized to capture the statistical dependencies amongst bus voltages. We prove that the bus connectivity and grid topology estimation problems, in radial and mesh structures, can be formulated as a linear regression with a least absolute shrinkage regularization on grouped variables (\textit{group lasso}). Simulations show highly accurate results in eight MV and LV distribution networks at different sizes and 22 topology configurations using PG\&E residential smart meter data.
\end{abstract}


%
\IEEEpeerreviewmaketitle

\section{Introduction}
A core mission of building Smart Cities is providing sustainable and economical energy. To achieve this goal, distributed energy resources (DERs), such as photovoltaic (PV) devices, energy storage devices, and electric vehicles, have been deeply integrated into the distribution grids to provide sustainable energy and reduce electricity cost. Such a trend will continue in the future deepening the DERs penetration further. 

While offering new opportunities, the increasing DER penetration triggers reliability risks to the operation of distribution systems. The distributed generation and the bidirectional power flow can cause the installed protective devices and operation systems to become insufficient. For example, the local grid may become unstable with the presence of even a small-scale of DERs \cite{dey2010urban}. Also, the voltage unbalance and transformer overload may occur due to the frequent plug-in electric vehicles in the low voltage grid \cite{clement2010impact}. For the future distribution grid with deeply penetrated DERs, better grid monitoring tools (for islanding and line work hazards) are needed for system operation, where topology information (for one or more new buses) is a prerequisite. 

In transmission grids, the grid topology is usually available to system operators. The topology error caused by the infrequent reconfiguration can be identified by the topology processor and state estimation\cite{huang2012electric, abur2004power, Lugtu80}. Unfortunately, in medium voltage (MV) and low voltage (LV) distribution grids, a topology can frequently change and make existing methods with limited performance. Furthermore, in many major metropolitan areas and industrial parks, MV/LV distribution grid branches are immense and mostly underground. For example, in New York City, over 94,000 miles of distribution lines are underground\cite{rudin2012machine}. Thus, the installation of topology identification devices in urban distribution grids is time-consuming and expensive. Even worse, the methods proposed for overhead transmission grid become infeasible in underground distribution grids due to the frequent topology reconfiguration \cite{brown2008impact}.

These critical challenges discussed above cause many previous assumptions on topology estimation to become invalid. For example,\cite{hayes2016event,arghandeh2015topology, baran2009topology} require the locations of switches or admittance matrices and look for the most likely topology from a collection of configurations. In \cite{deka2015structure} and \cite{sharon2012topology}, the impedances are also needed if only partial measurements are available. These requirements become unsuitable in metropolitan distribution systems because switch connectivity statuses and admittance matrices are often unavailable. Also, in many field applications, these details may be outdated or inaccurately recorded due to unreported power engineering activities, e.g., manual outage restoration in the substations. Furthermore, many DERs in distribution grids are not owned or operated by the utilities. Thus, their operation information may be inaccessible to the utilities. \cite{li2013blind} and \cite{anwar2016estimation} use the DC approximation and SCADA data to estimate the grid topology. However, the branches in distribution grid usually have non-negligible resistance. Furthermore, \cite{peppanen2016distribution,deka2016estimating,deka2016tractable} are designed for radial networks only, but many LV distribution grids are mesh in metropolitan districts and in regions with high load densities (e.g., industrial parks)  \cite{rudin2012machine,allan2013reliability,diaz2002application}. Many MV distribution grids have mesh structure and are operated with a radial topology. Several utilities, such as Taipower, Florida Power Company, Hong Kong Electric Company, Singapore Power, and Korea Electric Power Cooperation, have operated mesh (closed-loop) MV distribution grids in their service zones \cite{chen2004feasibility,kim2013advanced,jeon2016underground,pagel2000energizing,teo1995principles}. Recent studies \cite{celli2004meshed,de2014investigation} show that MV grid with mesh operational topology is more reliable and efficient with high penetration of DERs. \rev{\cite{cavraro2017voltage} proposes a maximum posterior probability approach to identify mesh operational topology from a candidate pool. In \cite{talukdar2017learning}, the distribution grid is formulated as a linear dynamic system and a Wiener filtering-based method is proposed to recover the radial and mesh structures.} At last, \cite{cavraro2015data,deka2016estimating,yuan2016inverse} require the phasor measurement units (PMUs), which are not widely available in current distribution grids.

The increasing investment in Advanced Metering Infrastructure (AMI) provides a new opportunity to utilize the historical data to solve new problems, such as estimating the underground distribution line connectivity in urban areas\cite{dey2010urban}. Hence, we only use household smart meter data in this paper, which include voltage magnitude, real power, and reactive power. Our goal is helping these buses to locate their connectivity to each other and the backbone local grid.

Mathematically, we firstly represent an MV/LV distribution system in a probabilistic graphical model. Then, we propose a method that estimates the connectivity of a bus. \rev{By exploiting the linear relationship between nodal voltages and injected currents, this algorithm uses the historical data of voltage phasors to fit a linear regression with the $L_1$ penalty on grouped variables, which is known as the \textit{group lasso} problem. Based on the bus connectivity estimation, we extend the group lasso approach to reconstruct network topology when multiple buses have uncertain connectedness. Furthermore, the voltage phase angle is usually unavailable in distribution grids due to the lack of PMU deployment. To address this issue, we utilize two approximations in distribution grids and extend the proposed algorithm to only use voltage magnitude to recover distribution grid topology.} Compared with existing approaches, our generalized grouping-based method shows several advantages. Firstly, our method can estimate a mesh grid with a limited amount of data. Secondly, our algorithm has no error propagation because the bus connectivity is estimated independently \cite{bolognani2013identification,liao2015distribution}. Thirdly, the computational complexity of the proposed algorithm is linear in term of data length. Finally, our approach has a reliable performance with noisy measurements.

Our data-driven algorithm is validated by the simulations of two IEEE distribution test cases \cite{teng2002modified,kersting2001radial} and six European MV and LV distribution grids \cite{pretticodistribution} with 18 network configurations. We also utilize Pacific Gas and Electric Company (PG\&E) residential smart meter measurements and emulated rooftop PV generation data \cite{dobos2014pvwatts} from National Renewable Energy Laboratory (NREL) for simulations. The numerical results show that our method outperforms recent works, especially in mesh systems \cite{bolognani2013identification,liao2015distribution}. Compared with our previous work \cite{liao2016urbanpes}, we firstly prove that the incremental change of voltage measurements can also be used to reconstruct the distribution grid topology. Secondly, unlike \cite{liao2016urbanpes}, we only apply the regularized linear regression on a subgroup of variables. Thirdly, we validate our algorithm on more network topologies and configurations with real data.

For the remainder of this paper, the MV/LV distribution grid model, its graphical model representation, and the data-driven topology estimation problem are presented in Section~\ref{sec:model}. We prove that the bus connectivity can be efficiently estimated by a linear regression with $L_1$ regularization in Section~\ref{sec:main}. In Section~\ref{sec:joint}, we formulate the grid topology reconstruction process as a convex optimization problem. Section~\ref{sec:num} validates the performance of proposed methods using multiple grids and real data.  Section~\ref{sec:con} gives the conclusions.

\section{MV/LV Distribution Grid Model}\label{sec:model}
An MV/LV distribution grid is composited by buses and branches. To embed the smart meter information into the topology estimation problem, for an $M$-bus system, we build a probabilistic graphical model $\mathcal{G}=(\mathcal{M},\mathcal{E})$ with a set of vertices $\mathcal{M}=\{1,2,\cdots,M\}$ and a set of unidirectional edges $\mathcal{E} = \{e_{ik},i,k \in \mathcal{M}\}$. In our graphical model $\mathcal{G}$, a vertex is represented by a random variable $V_i$ and corresponds to a bus. If the measurements at bus $i$ and $k$ exist a statistical dependence, then an edge $e_{ik}$ is in the edge set $\mathcal{E}$. The visualization of the distribution system and its graphical model representation are presented in Fig.~\ref{fig:cyber_physical}. For bus $i$, the voltage measurement at time $t$ is $v_i[t] = |v_i[t]|e^{j\theta_i[t]} \in \complex$. The units of the magnitude $|v_i[t]| \in \reals$ and phase angle $\theta_i[t] \in \reals$ are per unit and degree, respectively. Bus $1$ is assumed to be the slack bus with constant magnitude and phase angle. All measurements are noiseless and in the steady state. In Section~\ref{sec:num}, we will discuss the cases with noisy measurements.

\putFig{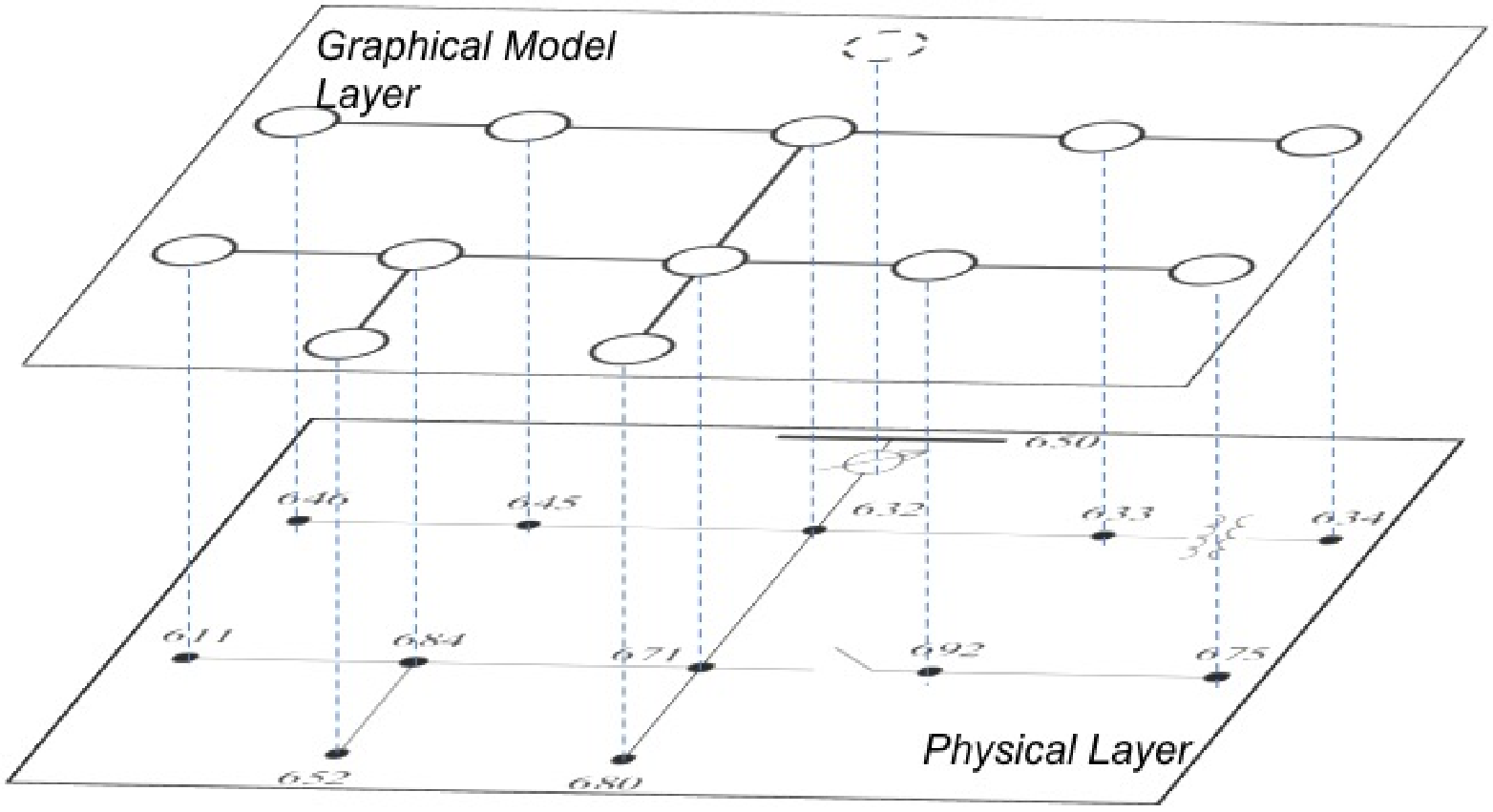}{The representation of a physical network and its corresponding graphical model $\mathcal{G}$.}{0.7\linewidth}
In many MV and LV distribution grids, voltage measurements have an irregular probability distribution. To better formulate the bus connectivity estimation problem, the incremental changes in the voltage measurements are used to estimate the grid topology \cite{chen2016quickest}. The incremental voltage change at bus $i$ is $\Delta v_i[t] = v_i[t] - v_i[t-1]$. $\Delta v_i[1] = 0$ when $t=1$. Since bus $1$ is the slack bus, $\Delta v_1[t] = 0$ for all $t$. $\Delta V_i$ represents the random variable of voltage change.

\putFig{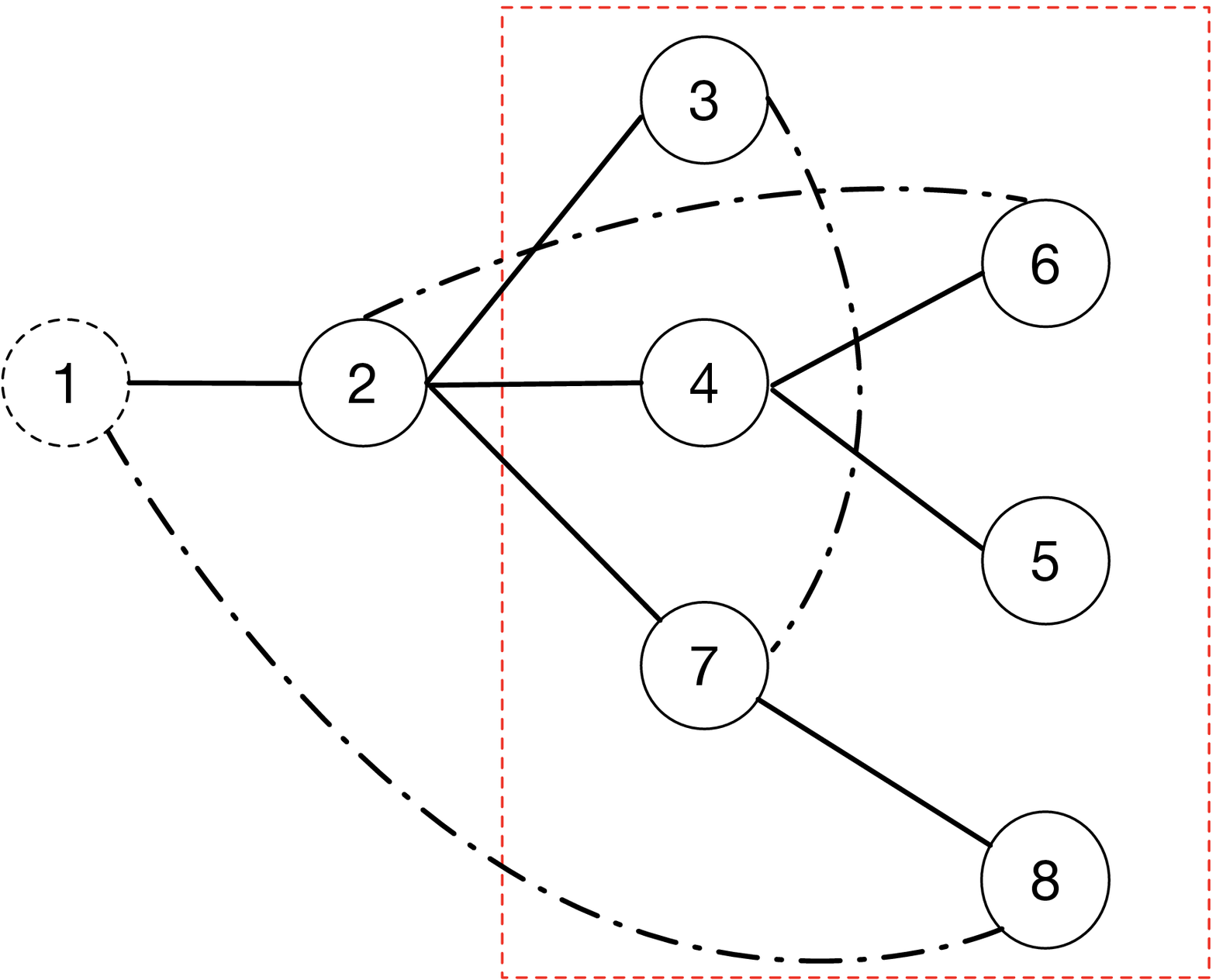}{A modified mesh $8$-bus system \cite{teng2002modified}. The bus connectivity within the red dashed box is unknown. The dashed branches are added to create mesh structures.}{0.6\linewidth}

With the system modeling above, the problem we want to address in this paper is defined as follows:
\begin{itemize}
\item Problem: data-driven bus connectivity and grid topology estimation based on bus voltage incremental changes
\item Given: the time-series voltage incremental measurements $\Delta v_i[t], t = 1,\cdots,T, i \in \mathcal{M}$ and a grid with partially known topology, as shown in Fig.~\ref{fig:new_8bus}
\item Find: (1) the bus connectivity; (2) the unknown grid topology $\mathcal{E}$.
\end{itemize}

\section{Bus Connectivity Estimation with Group Lasso}
\label{sec:main}
\subsection{Problem Formulation}
\label{sec:formulation}
In our graphical model, bus voltage incremental changes are modeled as random variables. Using chain rules, the joint probability $\P(\Delta \mathbf{V}_\mathcal{M})$ can be expressed as 
\begin{align}
&\P(\Delta V_2,\Delta V_3,\cdots,\Delta V_M) \nonumber \\
=& \P(\Delta V_2)\P(\Delta V_3|\Delta V_2)\cdots \P(\Delta V_M|\Delta V_2,\cdots,\Delta V_{M-1}). \label{eq:joint}
\end{align}
Bus 1 is omitted because it is the slack bus with constant voltage. If the slack bus does not have a constant voltage, we model the slack bus voltage incremental change as a random variable ($\Delta V_1$) and include it in (\ref{eq:joint}). In a distribution grid, adjacent buses are highly correlated \cite{liao2015distribution}. Therefore, we can approximate the joint probability $\P(\Delta \mathbf{V}_\mathcal{M})$ as
\begin{equation}
\label{eq:dist}
\P(\Delta \mathbf{V}_\mathcal{M}) \simeq \prod_{i=2}^M\P(\Delta V_i|\Delta\mathbf{V}_{\mathcal{N}(i)}),
\end{equation}
where $\mathcal{N}(i)$ denotes the neighbor set that includes the adjacent buses of bus $i$, i.e., $\mathcal{N}(i) = \{k \in \mathcal{M} | e_{ik}\in \mathcal{E}\}$. If this approximation holds, finding the bus connectivity is equivalent to finding the adjacent buses. The existing works \cite{peppanen2016distribution,deka2016estimating,deka2016tractable}, are restricted to only find parent nodes because of the radial topology assumption. However, an MV/LV distribution grid topology in urban area can be radial or mesh. Our goal is proposing a method that is suitable for both types. Next, we will take a two-stage proof to show  that $\Delta V_i$ only has statistical dependency with its adjacent buses under an appropriate assumption. We will also show why the approximation of $\P(\Delta\mathbf{V}_\mathcal{M})$ in (\ref{eq:dist}) holds. In the rest of this paper, the set complement, i.e., $\mathcal{X}\backslash \mathcal{Y} = \{i\in \mathcal{X}, i \notin \mathcal{Y}\}$, is represented by the operator $\backslash$.

\begin{assumption}
\label{ass:ass1}
In distribution grid, 
\begin{enumerate}
	\item the incremental change of the current injection $\Delta I$ at each non-slack bus is independent, i.e., $\Delta I_i \perp \Delta I_k$ for all $i \neq k$. 
	\item the incremental changes of the current injection $\DI$ and bus voltage $\DV$ follow Gaussian distribution with zero means and non-zero variances.
\end{enumerate}
\end{assumption}

Fig.~\ref{fig:IP_indept} illustrates the pairwise mutual information of incremental changes in bus current injection. The mutual information $I(X, Y)$ is a measure of the statistical dependence between two random variables $X$ and $Y$. When the mutual information is zero, these two random variables are independent, i.e., $X \perp Y$ \cite{cover2012elements}. In Fig.~\ref{fig:IP_indept}, the relatively small mutual information means that one can approximate that the current injections are independent with some approximation errors. This assumption has also been adopted in other works, such as \cite{deka2015structure,deka2017topology}. \rev{To further validate the independency of current injection increment $\Delta I$, we illustrate the average autocorrelation of current injection increment in \textit{LV\_suburban} system and IEEE 123-bus system. The error bar is one standard deviation. We can observe that in both LV and MV distribution grids, the autocorrelation of $\Delta I$ drops significantly as the lag increases. This observation proves that the current injection increment is independent over time.}

\putFig{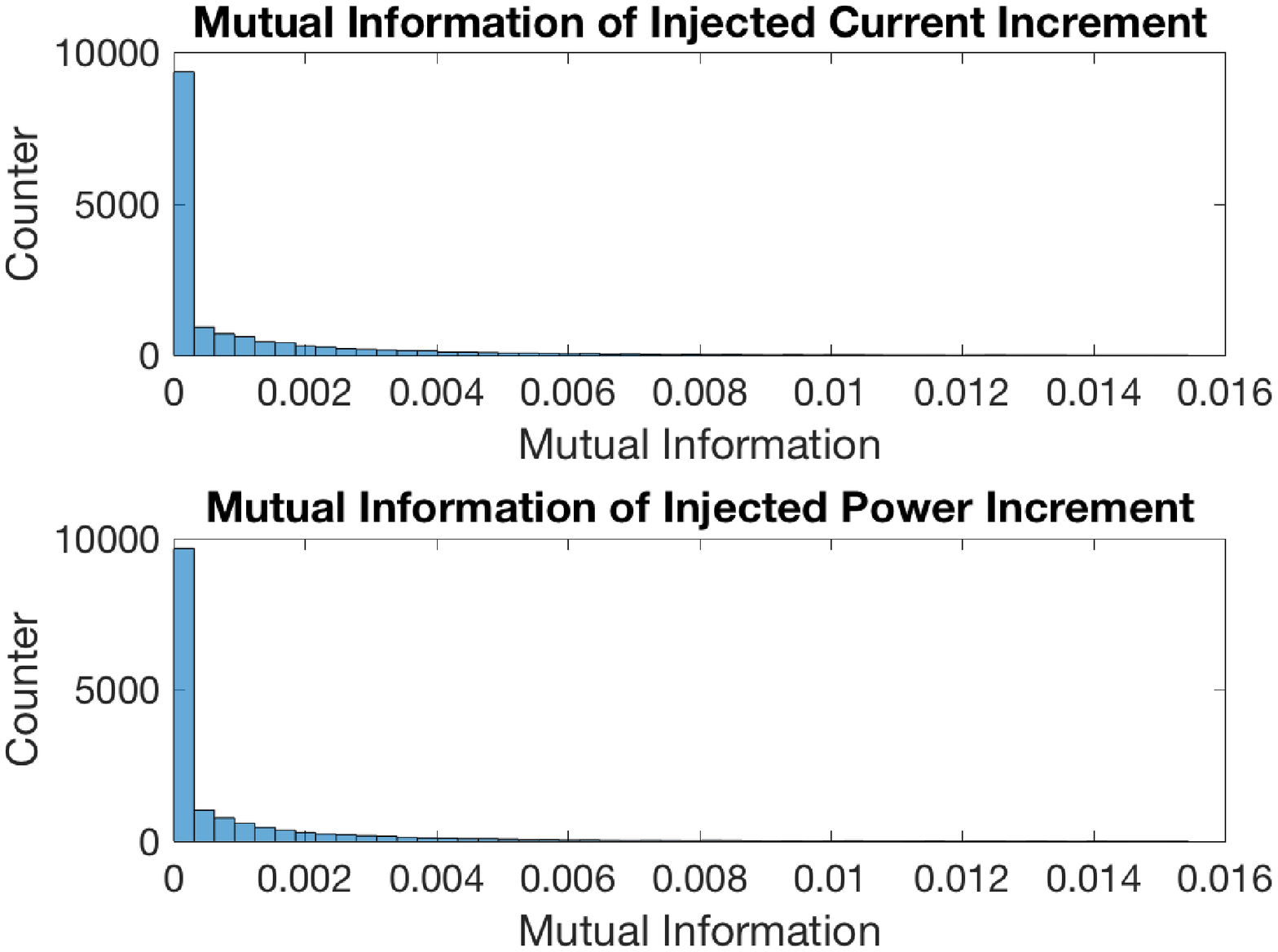}{\rev{Mutual information of pairwise current injection increment $\Delta I$ and power injection increment $\Delta P$ in the IEEE 123-bus system.}}{0.9\linewidth}

\putFig{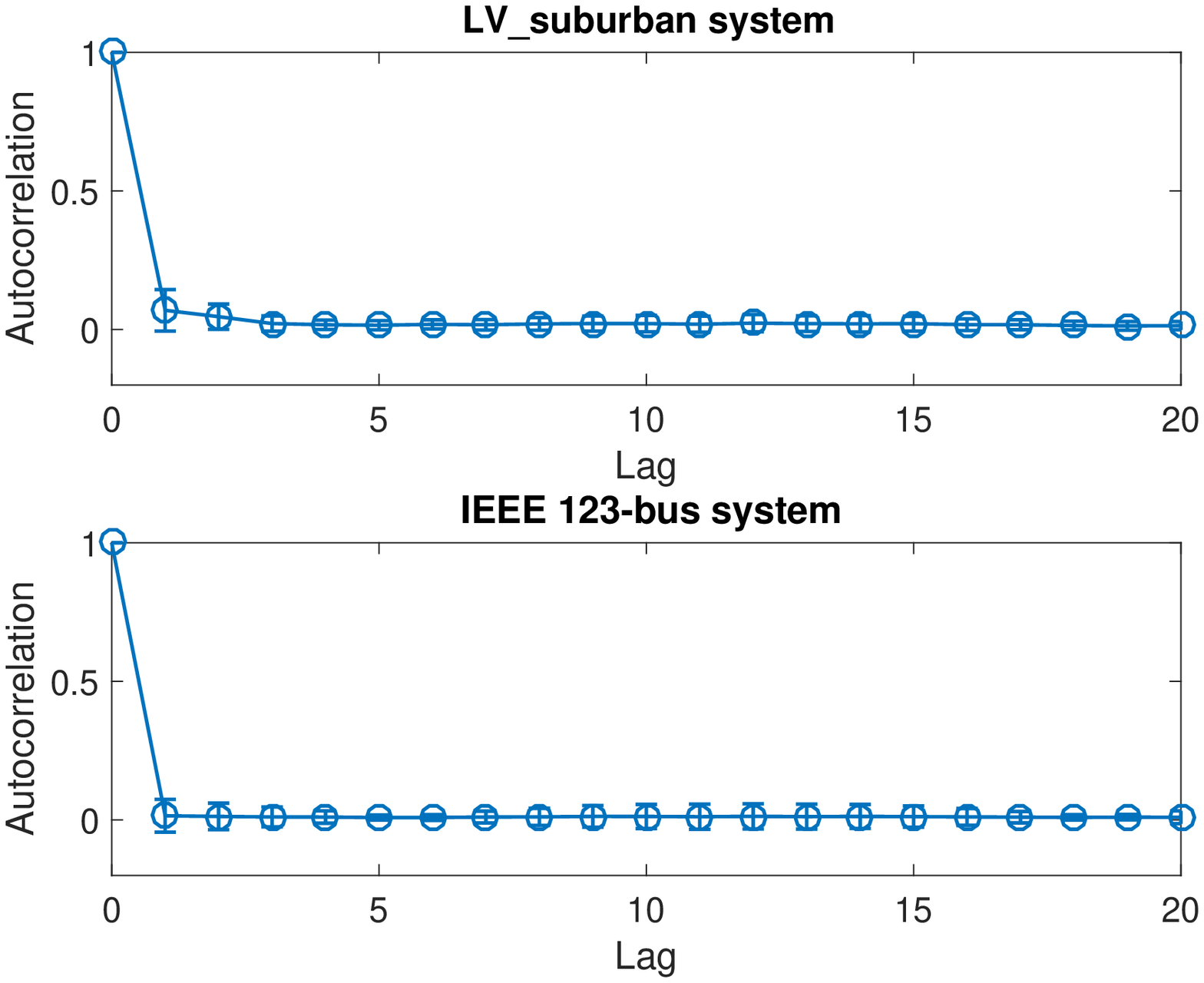}{\rev{Average autocorrelation of current injection increment $\Delta I$ of \textit{LV\_suburban} system and IEEE 123-bus system. The error bar is one standard deviation.}}{0.9\linewidth}

\rev{
\begin{remark}
In some distribution grid topology estimation works, the injected power increment independence is adopted, instead of injected current increment independence. In distribution grid, the end-user load models depend on many factors, such as load type, time frame, and voltage balance \cite{moller2016probabilistic,collin2014development,schneider2011multi}. In Fig.~\ref{fig:IP_indept}, we also illustrate the mutual information of pairwise power injection increment $\Delta P$ in the IEEE 123-bus system. The histograms of mutual information of $\Delta I$ and $\Delta P$ are similar. Therefore, for the data sets we used in this paper, both independence assumptions are held. We prefer the assumption of current injection independence because it simplifies the proof of following theorems and lemmas.
\end{remark}
}

As the first stage of proof, we will show that $\P(\Delta \mathbf{V}_\mathcal{M}) = \prod_{i=2}^M\P(\Delta V_i|\Delta\mathbf{V}_{\{\mathcal{N}(i)\cup \mathcal{N}^\RN{2}(i)\}})$, where $\mathcal{N}^\RN{2}(i)$ denotes the set that includes the adjacent buses' indices of bus $i$'s neighbors (buses that are two-hop away from bus $i$), i.e., $\mathcal{N}^\RN{2}(i) = \{k \in \mathcal{M}|e_{kl} \in \mathcal{E}, l \in \mathcal{N}(i)\}$.

\rev{
\begin{lemma}
\label{lemma: linear_indept}
	Let $X_1$, $X_2$, and $Y$ be Gaussian random variables, where $X_1$ and $X_2$ are independent. Then, given the following equation, 
	\[
		a_1 X_1 + b_1 X_2 = Y
	\]
	where $a_1, b_1\in \reals$, $X_1$ and $X_2$ are conditionally independent given $Y=y$ if $a_1 = 0$, $b_1 = 0$, or $a_1 = b_1 = 0$. (see Appendix~\ref{sec:lemma1_proof} for proof.)
\end{lemma}

\begin{remark}
	When $\mathbf{X}_1,\mathbf{X}_2$, and $\mathbf{Y}$,  are Gaussian random vectors, Lemma~\ref{lemma: linear_indept} can  be extended to prove that $\mathbf{a}_1^T\mathbf{b}_1 = \mathbf{0}$ is the necessary condition that $\mathbf{X}_1 \perp \mathbf{X}_2 | \mathbf{Y}=\mathbf{y}$, where $\mathbf{a}_1^T\mathbf{X}_1 + \mathbf{b}_1^T\mathbf{X}_2 = \mathbf{Y}$.
\end{remark}
}

\rev{
\begin{lemma}
\label{lemma:linear_indept2}
	Let $X_1$, $X_2$, $Y$, $Z$ be Gaussian random variables, where $X_1$ and $X_2$ are independent. Then, given the following equations,
	\begin{eqnarray}
		c_1 Y + d_1 Z &=& X_1 \label{eq:lemma2_1}\\
		c_2 Y + d_2 Z &=& X_2,\label{eq:lemma2_2}
	\end{eqnarray}
	where $c_1, d_1, c_2, d_2 \in \reals$, $X_1$ and $X_2$ are conditionally independent given $Z=z$ if $c_1 = 0$ or $c_2 = 0$. (see Appendix~\ref{sec:lemma2_proof} for proof.)
\end{lemma}

\begin{remark}
	When $\mathbf{X}_1,\mathbf{X}_2,\mathbf{Y}, \mathbf{Z}$ are Gaussian random vectors, Lemma~\ref{lemma:linear_indept2} can be extended to prove that $\mathbf{c}_1^T\mathbf{c}_2 = \mathbf{0}$ and $\mathbf{c}_1 \neq \mathbf{c}_2 \neq 0$ are the necessary condition that $\mathbf{X}_1 \perp \mathbf{X}_2 | \mathbf{Z}=\mathbf{z}$, where $\mathbf{c}_1^T\mathbf{Y} + \mathbf{d}_1^T\mathbf{Z} = \mathbf{X}_1$ and $\mathbf{c}_2^T\mathbf{Y} + \mathbf{d}_2^T\mathbf{Z} = \mathbf{X}_2$.
\end{remark}
}

\begin{theorem}
\label{thm:two_step_dependency}
	In a distribution grid, the incremental voltage change of bus $i$ and the incremental voltage changes of all other buses that are not in $\{\mathcal{N}(i) \cup \mathcal{N}^\RN{2}(i)\}$ are conditionally independent, given the incremental voltage changes of buses in $\{\mathcal{N}(i) \cup \mathcal{N}^\RN{2}(i)\}$, i.e., $\Delta V_i \perp \left\{\Delta V_k, k \notin \{i,\mathcal{N}(i), \mathcal{N}^\RN{2}(i)\} \right\} |\{\Delta V_q, q \in \mathcal{N}(i) \cup \mathcal{N}^\RN{2}(i)\}$.
\end{theorem}
\putFig{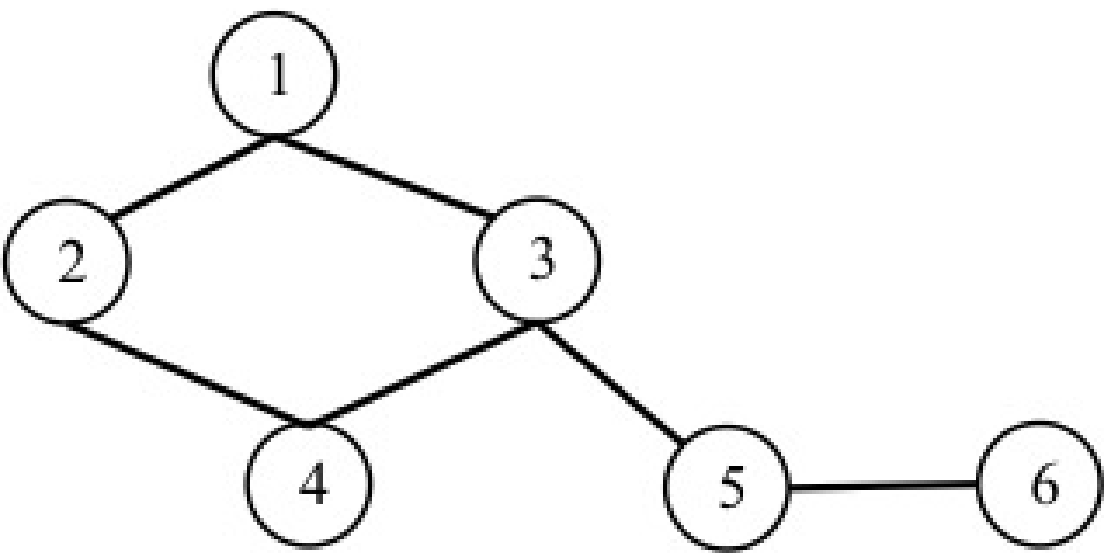}{An example used to show Theorem~\ref{thm:two_step_dependency}.}{0.45\linewidth}

We will use a simple example to show the conditional independence. A formal proof is given in Appendix~\ref{sec:thm1_proof}. Using the circuit equation $\mathbf{Y}\Delta\mathbf{V} = \Delta\mathbf{I}$, the system in Fig.~\ref{fig:6bus_loop} is expressed as:
\begin{equation*}
\thickmuskip=-2mu
\begin{bmatrix}
y_{11} & -y_{12} & -y_{13} & 0 & 0 & 0\\
-y_{12} & y_{22} & 0 & -y_{24} & 0 & 0\\
-y_{13} & 0 & y_{33} & -y_{34} & -y_{35} & 0 \\
0 & -y_{24} & -y_{34} & y_{44} & 0 & 0\\
0 & 0 & -y_{35} & 0 & y_{55} & -y_{56} \\
0 & 0 & 0 & 0 & -y_{56} & y_{66}
\end{bmatrix}
\begin{bmatrix}
\Delta V_1 \\ \Delta V_2 \\ \Delta V_3 \\ \Delta V_4 \\ \Delta V_5 \\ \Delta V_6
\end{bmatrix}
=
\begin{bmatrix}
\Delta I_1 \\ \Delta I_2 \\ \Delta I_3 \\ \Delta I_4 \\ \Delta I_5 \\ \Delta I_6
\end{bmatrix}
\end{equation*}
where $y_{ik} = y_{ki}$ denotes the deterministic admittance between bus $i$ and $k$, $y_{ii} = \sum_{k=1,i\neq k}^6y_{ik}+\frac{1}{2}b_i$ for $i = 2,\cdots, 6$, and $b_i$ is the shunt admittance at bus $i$. For bus 1, which connects with the slack bus, $y_{11} = y_{01}+\sum_{k=1,i\neq k}^6y_{ik}+\frac{1}{2}b_i$, where $y_{01} \neq 0$ is the admittance of the branch that connects bus $1$ and the slack bus $0$. If $y_{ik} = 0$, there is no branch between bus $i$ and $k$.

For bus $2$, the neighbor set $\mathcal{N}(2) = \{1,4\}$ and the two-hop neighbor set $\mathcal{N}^\RN{2}(2) = \{3\}$. Given $\Delta V_1 = \Delta v_1$, $\Delta V_3 = \Delta v_3$, and $\Delta V_4 = \Delta v_4$, we have following equations:
\begin{eqnarray}
\DI_1 &=& y_{11}\Dv_1 - y_{12}\DV_2 - y_{13}\Dv_3 \label{eq:v1}\\
\DI_2 &=& -y_{12}\Dv_1 + y_{22}\DV_2 - y_{24}\Dv_4 \label{eq:v2} \\
\DI_3 &=& -y_{13}\Dv_1 + y_{33}\Dv_3 - y_{34}\Dv_4 - y_{35}\DV_5 \label{eq:v3}\\
\DI_4 &=& -y_{24}\DV_2 - y_{34}\Dv_3 + y_{44}\Dv_4 \label{eq:v4} \\
\DI_5 &=& -y_{35}\Dv_3  + y_{55}\DV_5 - y_{56}\DV_6 \label{eq:v5}\\
\DI_6 &=& -y_{56}\DV_5 + y_{66}\DV_6 \label{eq:v6}
\end{eqnarray}

To prove the conditional independency of $\DV$, we firstly need to check if the independency among $\DI$ still holds, given $\DV_1, \DV_3$ and $\DV_4$. Let's consider $X_1 = \DI_2, X_2 = \DI_3$, $\mathbf{Y} = [\DV_2,\DV_5]$, and $\mathbf{Z} = [\DV_1,\DV_3,\DV_4]$. Using Lemma~\ref{lemma:linear_indept2}, we know that $\DI_2 \perp \DI_3$ given $\{\DV_1,\DV_3,\DV_4\}$. Therefore, $\DV_2$ and $\DV_5$ are conditionally independent, according to (\ref{eq:v2}) and (\ref{eq:v3}). 

To prove the conditional independence between $\DV_2$ and $\DV_6$, we combine (\ref{eq:v3}) and (\ref{eq:v6}) and have the following equation:
\[
\DI_6 - \frac{y_{56}}{y_{35}}\DI_3 = \frac{y_{56}}{y_{35}}(y_{13}\Dv_1 - y_{33}\Dv_3 + y_{34}\Dv_4) + y_{66}\DV_6.
\]
Applying Lemma~\ref{lemma:linear_indept2}, we prove that $\DI_2$ and $\DI_3+\DI_6$ are conditionally independent, given $\{\DV_1,\DV_3,\DV_4\}$. Therefore, $\DV_2$ and $\DV_6$ are conditionally independent given $\{\DV_1,\DV_3,\DV_4\}$. We can extend this approach to other pairs of buses and prove that Theorem~\ref{thm:two_step_dependency} holds for this example. 

\putFig{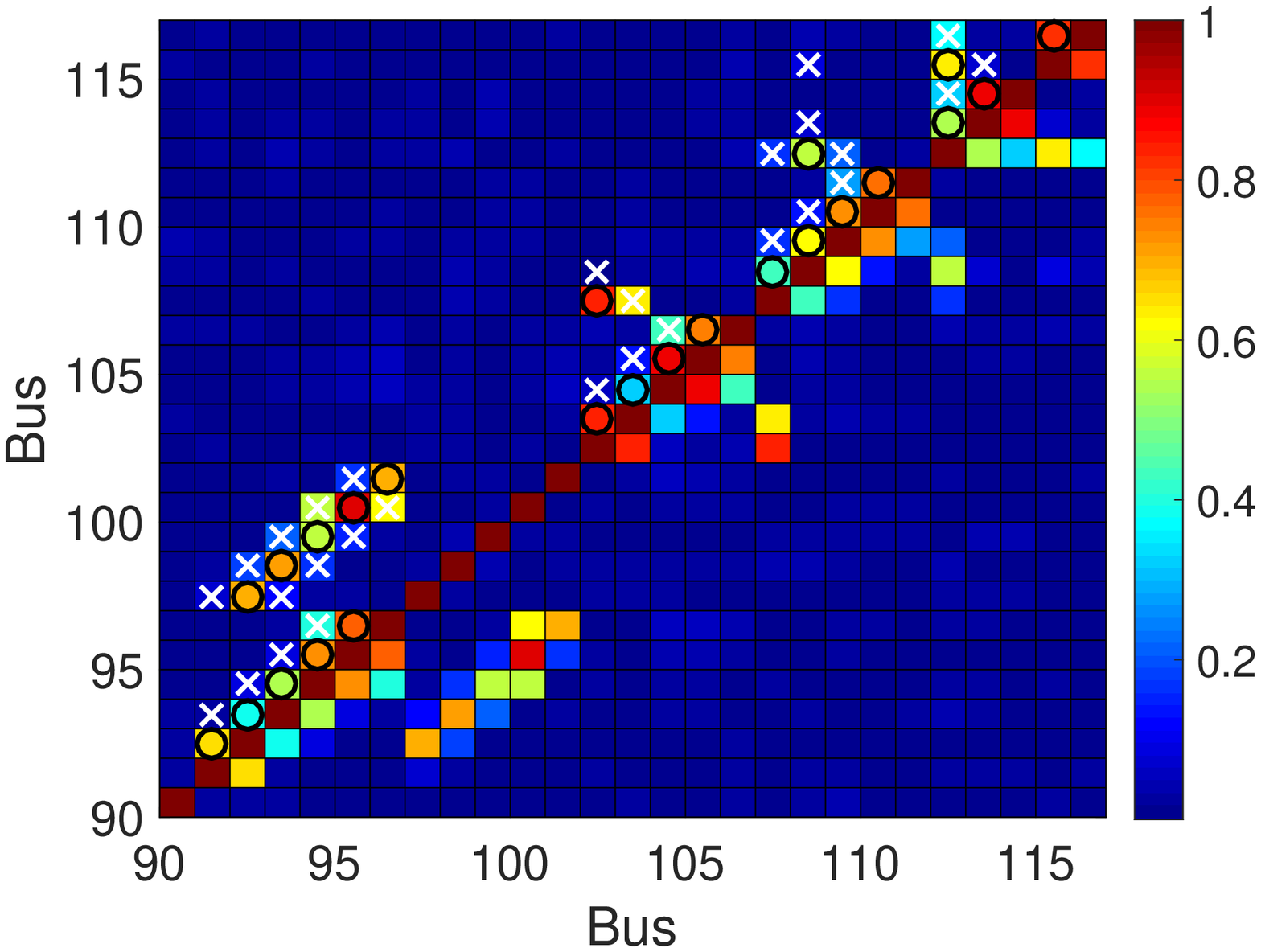}{\rev{Conditional correlation between buses in IEEE 123-bus system. The circle indicates the neighbors of bus $i$. The crossing indicates the two-hop neighbor of bus $i$. The square without markers represents the bus pair that are more than two-hop away.}}{\linewidth}

With Theorem~\ref{thm:two_step_dependency}, we can show that $\P(\Delta \mathbf{V}_\mathcal{M}) = \prod_{i=2}^M\P(\Delta V_i|\Delta\mathbf{V}_{\mathcal{N}(i)\cup \mathcal{N}^\RN{2}(i)})$. This observation is similar to the results in \cite{deka2017topology}. 

\rev{In Fig.~\ref{fig:bus_corr}, we show the conditional correlations of voltage increments between each bus pair in IEEE 123-bus distribution system using the real load data from PG\&E. The distribution grid configuration and simulation setup are described in Section~\ref{sec:num}. In Fig.~\ref{fig:bus_corr}, the color in a square represents the absolute conditional correlation of voltage increments of two buses. As discussed in \cite{hastie2015statistical}, if the voltage increments of two buses are independent, their conditional correlation is zero (dark color). In Fig.~\ref{fig:bus_corr}, the circle refers to the bus neighbors and the crossing indicates the two-hop neighbors. If a square without any marker, it means the pair of buses is more than two-hop away. As illustrated in Fig.~\ref{fig:bus_corr}, the conditional correlation between the voltages of two-hop neighbors is higher than the conditional correlations of other bus pairs, but it is still much lower than the conditional correlation between two neighbors. The diagonal bus pairs have conditional correlation of 1 because it is the self correlation. Based on this observation, we make the following assumption:}
\begin{assumption}
\label{ass:ass2}
	In a distribution grid, given $\DV_{\mathcal{N}(i)}$, the conditional correlations between $\DV_i$ and $\Delta \mathbf{V}_{\mathcal{N}^\RN{2}(i)}$ are relatively small.
\end{assumption}

With Assumption~\ref{ass:ass2}, we can simplify $\P(\Delta \mathbf{V}_{\mathcal{M}})$ to depend on the voltages of neighbors. \rev{As a highlight, unlike existing methods in \cite{yuan2016inverse,deka2017topology}, the assumption of our method is inspiring by real data observation. In Section~\ref{sec:num}, we use numerical simulation to demonstrate that this approximation does not degrade the performance of topology estimation.}

\begin{lemma}
\label{lemma:one_step_cond_indept}
In a distribution grid, the voltage change of bus $i$ and the voltage changes of all other buses that are not connected with bus $i$ are conditionally independent, given the voltage changes of the neighbors of bus $i$, i.e., $\Delta V_i \perp \left\{\Delta V_k, k \in \mathcal{M}\backslash \{\mathcal{N}(i),i\} \right\}|\Delta \mathbf{V}_{\mathcal{N}(i)}$.
\end{lemma}


With Lemma~\ref{lemma:one_step_cond_indept} as the second stage of the proof, (\ref{eq:dist}) holds with equality, i.e., $\P(\Delta \mathbf{V}_\mathcal{M}) = \prod_{i=2}^M\P(\Delta V_i |\Delta\mathbf{V}_{\mathcal{N}(i)})$. Therefore, the voltage incremental change at each bus only depends on $\Delta V$ of its neighbors. In next subsections, we will propose how to find $\mathcal{N}(i)$ using $\Delta V$. Also, we will show the robustness where only $\Delta |V|$ is available.

\subsection{Bus Connectivity Reconstruction via Linear Regression}

\putFig{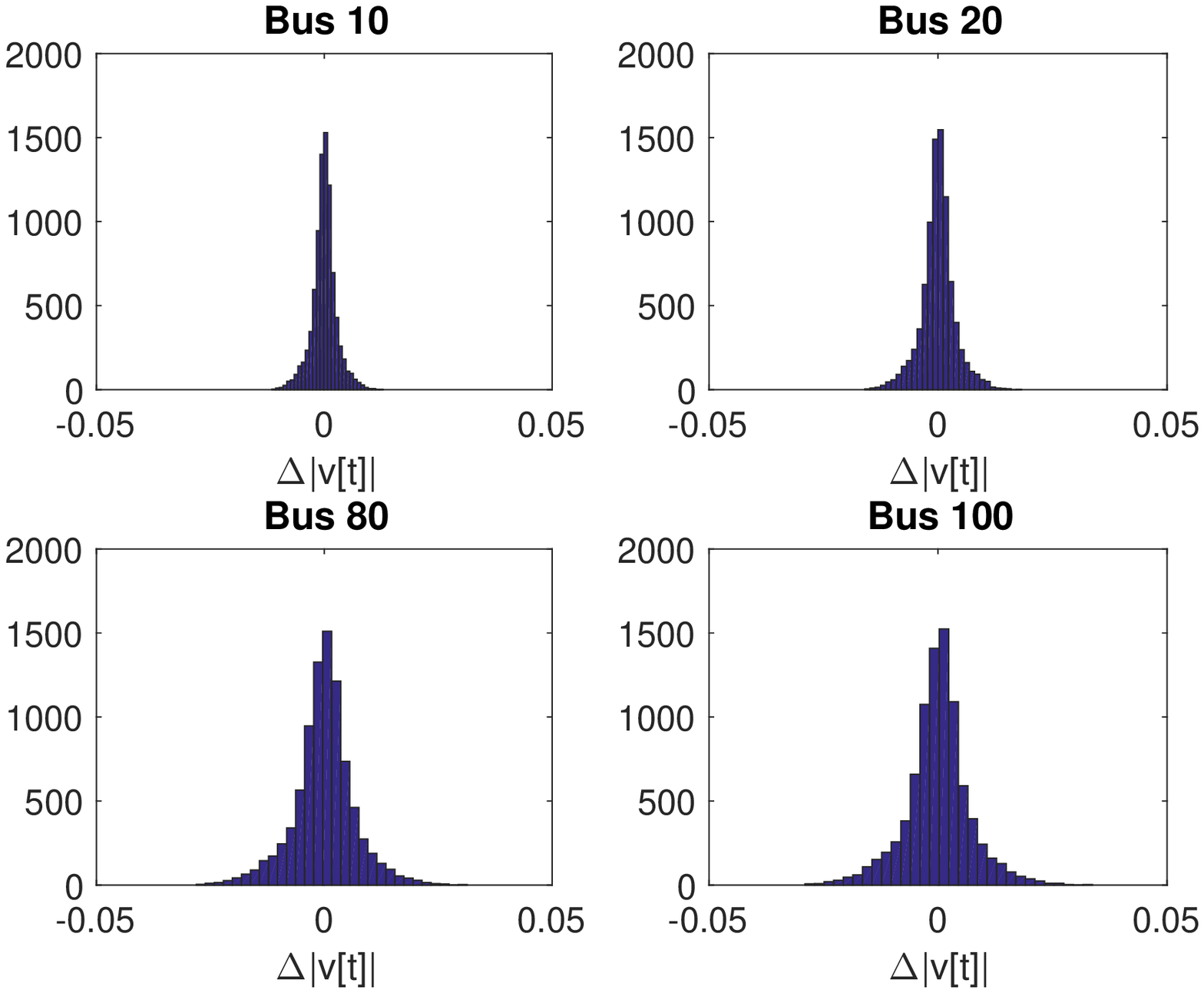}{Histogram of $\Delta|v[t]|$ of four buses in IEEE 123-bus system.}{0.8\linewidth}
For bus $i$, $\Delta \mathbf{V}_{\mathcal{M}\backslash\{i\}}$ is a collection of all nodal voltages in the graphical model $G$ beside $V_i$. As discussed in Assumption~\ref{ass:ass1}, we assume $\Delta\mathbf{V}_\mathcal{M}$ to be a Gaussian random vector, which has been empirically shown in Fig.~\ref{fig:dv}. The conditional distribution of $\Delta V_i$ given $\Delta \mathbf{V}_{\mathcal{M}\backslash\{i\}}$ is also a Gaussian random variable. Therefore, based on the probability density function of Gaussian distribution, $\Delta V_i$ is a linear equation of $\Delta \mathbf{V}_{\mathcal{M}\backslash\{i\}}$ and an error term $\epsilon_{\mathcal{M}\backslash\{i\}}$, i.e.,
\begin{equation}
\label{eq:lin_reg}
\Delta V_i = \Delta \mathbf{V}_{\mathcal{M}\backslash\{i\}}^H\boldsymbol{\beta}^{(i)} +  \epsilon_{\mathcal{M}\backslash\{i\}},
\end{equation}
where $\boldsymbol{\beta}^{(i)}$ denotes the parameter vector and $H$ denotes the transpose operator. The error term $\epsilon_{\mathcal{M}\backslash\{i\}}$ is a Gaussian variable with a zero mean and a variance of $\text{Var}(\Delta V_i | \Delta\mathbf{V}_{\mathcal{M}\backslash\{i\}})$. It is also independent with $\Delta \mathbf{V}_{\mathcal{M}\backslash\{i\}}$ \cite{meinshausen2006high}. Because of Lemma~\ref{lemma:one_step_cond_indept}, we know that $\Delta V_i$ and $\Delta V_k$ are conditionally dependent if there is an edge $e_{ik}$. The non-zero coefficient in $\boldsymbol{\beta}^{(i)}$ indicates that two nodes are  statistically correlated. Hence, the bus adjacency identification problem is equivalent to a linear regression problem. We can use the parameter estimate $\widehat{\boldsymbol{\beta}}^{(i)}$ to find the neighbors of bus $i$.

From a physical perspective, we can also show that the nonzero coefficients in the parameter vector $\boldsymbol{\beta}^{(i)}$ indicate the bus connectivity. Specifically, at bus $i$, the increments of current injection and nodal voltages have the following relationship:
\begin{eqnarray}
\Delta I_i &=& \Delta V_iy_{ii} - \sum_{k \in \mathcal{N}(i)}\Delta V_ky_{ik}, \nonumber \\
\Delta V_i &=&  \sum_{k \in \mathcal{N}(i)} \frac{y_{ik}}{y_{ii}}\Delta V_k + \frac{\Delta I_i}{y_{ii}}.\label{eq:v_linear}
\end{eqnarray} 
with $y_{ii} = \sum_{k \in \mathcal{N}(i)}y_{ik} +\frac{1}{2}b_i$. Compared with (\ref{eq:lin_reg}), we find that if bus $k$ connects with bus $i$, i.e., $k \in \mathcal{N}(i)$, the $k$-th element of $\boldsymbol{\beta}^{(i)}$ is $y_{ik}/y_{ii}$. If bus $d$ is not an element of the set $\mathcal{N}(i)$, the $d$-th element of $\boldsymbol{\beta}^{(i)}$ is zero. The reason is that $y_{id} = y_{di} = 0$ and the voltage changes are conditionally independent, as Lemma~\ref{lemma:one_step_cond_indept}. The variation introduced by $\Delta I_i$ is captured by $\epsilon_{\mathcal{M}\backslash\{i\}}$. If we assume $\Delta V$ has a zero mean, $\epsilon_{\mathcal{M}\backslash\{i\}}$ also has a zero mean and follows a Gaussian distribution. These results are consistent with our previous discussion. \rev{In some cases, $\Delta I_i$ may be correlated with $\sum \DV_k$. But in our simulation in Section~\ref{sec:num}, we find that the variation of $\Delta I_i$ is much smaller than the variation of $\sum\DV_k$. Hence, we approximate $\Delta I_i$ as the noise term of linear regression in (\ref{eq:lin_reg}).}

Many distribution grids are not fully connected. The graphical model $\mathcal{G}$ is sparse and many elements in $\boldsymbol{\beta}^{(i)}$ are zero. A well-known regularization to ensure the sparsity in a linear regression is $L_1$ norm. This formulation is known as \textit{Lasso} \cite{tibshirani1996regression}. In lasso, the objective function is the sum of squared errors with a constraint on the sum of the absolute values of parameters ($L_1$ norm), i.e.:
\begin{equation}
\label{eq:Lasso}
\widehat{\boldsymbol{\beta}}^{(i)} = \arg\min_{\boldsymbol{\beta}^{(i)}}  \sum_{t=1}^T\left(\Delta v_i[t]-\sum_{\substack{k=2 \\ k \neq i}}^M \Delta v_k[t]\boldsymbol{\beta}^{(i)}_k\right)^2 + \lambda\|\boldsymbol{\beta}^{(i)}\|_1,
\end{equation}
where the regularization parameter $\lambda$ is non-negative, $\|\boldsymbol{\beta}^{(i)}\|_1$ denotes the regularization term, and $\|.\|_1$ denotes $L_1$ norm. If $\lambda = 0$, (\ref{eq:Lasso}) becomes an ordinary least squares problem. The objective function of (\ref{eq:Lasso}) is convex and can be solved by many well-known methods \cite{efron2004least,friedman2010regularization}. When solving the lasso problem in (\ref{eq:Lasso}), the bus connectivity $\widehat{\mathcal{N}}(i)$ can be estimated by finding the indices of non-zero elements of $\widehat{\boldsymbol{\beta}}^{(i)}$. How to choose $\lambda$ is discussed in Section~\ref{sec:lambda_path}.

\subsection{Bus Neighbors Estimation via Group Lasso}
\label{sec:group_lasso}
In the previous section, we have formulated the bus connectivity estimation problem as a lasso problem. However, this formulation is difficult to solve by utilizing many well-known lasso solvers because these approaches only solve lasso problem with real numbers. In power systems, voltage and admittance are complex numbers. To address this issue, we propose a group lasso approach that converts a complex number lasso formulation to a real number lasso problem. 

For two arbitrary complex numbers $x$ and $y$, their product $z = xy$ is expressed as
\begin{eqnarray*}
	\re(z) &=& \re(x)\re(y) - \im(x)\im(y), \\
	\im(z) &=& \re(x)\im(y) + \im(x)\re(y).
\end{eqnarray*}
Thus, the linear equation in (\ref{eq:lin_reg}) can be rewritten as
\begin{eqnarray}
	\begin{bmatrix}
		\re(\Delta V_i) \\
		\im(\Delta V_i)
	\end{bmatrix}
 	&=& \sum_{\substack{k=2 \\ k \neq i}}^M
 	\begin{bmatrix}
		\re(\Delta V_k) & -\im(\Delta V_k) \\
		\im(\Delta V_k) & \re(\Delta V_k)
	\end{bmatrix}
	\begin{bmatrix}
		\re(\boldsymbol{\beta}^{(i)}_k) \\
		\im(\boldsymbol{\beta}^{(i)}_k) \\
	\end{bmatrix}, \nonumber \\
	\mathbf{Z}_i &=& \sum_{\substack{k=2 \\ k \neq i}}^M \mathbf{X}_k\boldsymbol{\gamma}^{(i)}_k =\mathbf{X}\boldsymbol{\gamma}^{(i)}\label{eq:group_lin}.
\end{eqnarray}
In (\ref{eq:group_lin}), we transform a complex linear equation to a real linear equation. We can apply the $L_1$ constraint to the new parameter vector $\boldsymbol{\gamma}^{(i)}$, which becomes an ordinary lasso problem. 

Solving the linear regression in (\ref{eq:group_lin}) with $L_1$ penalty $\|\boldsymbol{\gamma}^{(i)}\|_1$ results a sparse estimate $\widehat{\boldsymbol{\gamma}}^{(i)}$. However, we cannot guarantee that both $\re(\boldsymbol{\beta}^{(i)}_k)$ and $\im(\boldsymbol{\beta}^{(i)}_k)$ are zero or nonzero at the same time. If bus $k$ is not connected with bus $i$, $\boldsymbol{\beta}^{(i)}_k$ is zero in (\ref{eq:Lasso}). Thus, both $\re(\boldsymbol{\beta}^{(i)}_k)$ and $\im(\boldsymbol{\beta}^{(i)}_k)$ are zeros. To enforce the sparsity on both real and imaginary parts of $\boldsymbol{\beta}$, in (\ref{eq:group_lin}), we need to apply sparsity constraint to a group of elements in $\boldsymbol{\gamma}^{(i)}$ such that all elements within a group will be zero if one of them is zero. This problem formulation is known as \textit{Group Lasso} \cite{yuan2006model}. Particularly, we can estimate $\boldsymbol{\gamma}^{(i)}$ as follows:
\begin{equation}
\label{eq:group_lasso}
\widehat{\boldsymbol{\gamma}}^{(i)} = \arg\min_{\boldsymbol{\gamma}^{(i)}} \sum_{t=1}^T \left\|\mathbf{z}_i[t] - \sum_{\substack{k=2 \\ k \neq i}}^M \mathbf{x}_k[t]\boldsymbol{\gamma}^{(i)}_k\right\|_2^2 + \lambda \sum_{\substack{k=2 \\ k \neq i}}^M \|\boldsymbol{\gamma}^{(i)}_k\|_2.	
\end{equation}
Unlike the lasso formulation in (\ref{eq:Lasso}), in  (\ref{eq:group_lasso}), we use $L_2$ norm because it enforces the entire vector $\boldsymbol{\gamma}_k^{(i)}$ to be zero or nonzero. See \cite{hastie2015statistical} for more details on group lasso.

We can construct $\widehat{\boldsymbol{\beta}}^{(i)}$ from $\widehat{\boldsymbol{\gamma}}^{(i)}$ and find the non-zero elements in $\widehat{\boldsymbol{\beta}}^{(i)}$. Alternatively, if bus $k$ is not a neighbor of bus $i$, both elements in $\widehat{\boldsymbol{\gamma}}^{(i)}$ are zero. Algorithm~\ref{alg:alg1} summarizes the steps of proposed bus connectivity estimation algorithm.


\begin{algorithm}[h!]
\caption{Distribution Grid Bus Connectivity Estimation}
\begin{algorithmic}[1]
\REQUIRE $\Delta v_i[t]$ for $t = 1,\cdots,T$
\STATE For bus $i$, solve the group lasso problem in (\ref{eq:group_lasso}) and estimate the parameter vector $\widehat{\boldsymbol{\gamma}}^{(i)}$.
\STATE Compute $\widehat{\boldsymbol{\beta}}^{(i)}$ as $\widehat{\boldsymbol{\beta}}^{(i)}_k = \widehat{\boldsymbol{\gamma}}^{(i)}_{k,1} + j\widehat{\boldsymbol{\gamma}}^{(i)}_{k,2}$. 
\STATE Find $\widehat{\mathcal{N}}(i)$ as $\widehat{\mathcal{N}}(i) = \{k|\widehat{\boldsymbol{\beta}}^{(i)}_k \neq 0\}$
\end{algorithmic}
\label{alg:alg1}
\end{algorithm}

\subsection{Bus Connectivity Estimation using Voltage Magnitude Only}
In distribution grids, voltage angles $\theta$ are hard to acquire because PMUs are not widely available. When only the change of voltage magnitude $\Delta|V_i|$ is available, $\mathbf{X}_k$ and $\mathbf{Z}_i$ become $\Delta |V_k|$ and $\Delta |V_i|$ respectively. Also, $\boldsymbol{\gamma}_k^{(i)}$ reduces to a scalar. The objective function of group lasso problem in (\ref{eq:group_lasso}) becomes 
\begin{eqnarray}
	&& \sum_{t=1}^T \left(\Delta | v_i[t]| - \sum_{\substack{k=2\\k \neq i}}^M \Delta | v_k[t]|\gamma^{(i)}_k\right)^2 + \lambda \sum_{\substack{k=2 \\ k \neq i}}^M \|\gamma^{(i)}_k\|_2 \nonumber \\
	&=& \sum_{t=1}^T \left(\Delta | v_i[t]| - \sum_{\substack{k=2 \\ k \neq i}}^M \Delta | v_k[t]|\gamma^{(i)}_k\right)^2 + \lambda \|\boldsymbol{\gamma}^{(i)}\|_1, \label{eq:V_lasso}
\end{eqnarray}
where $\Delta |v_i[t]| = |v_i[t]| - |v_i[t-1]|$ and $\Delta |v_i[1]| = 0$. In (\ref{eq:V_lasso}), $\|\gamma^{(i)}_k\|_2$ is equivalent to $|\gamma^{(i)}_k|$ and $\sum_k \|\gamma^{(i)}_k\|_2 = \sum_k |\gamma^{(i)}_k| = \|\boldsymbol{\gamma}^{(i)}\|_1$, where $\boldsymbol{\gamma}^{(i)} \in \reals^{M-2}$. Hence, with $\Delta|V|$ only, we can reconstruct the bus connectivity using the ordinary lasso. \rev{Unlike transmission grid, the $R/X$ ratio is large in distribution grids. Because of non-negligible branch resistance, the voltage measurements have larger variation. As proved in \cite{weng2016distributed}, in distribution grid, the statical correlations among bus voltage magnitudes are more significant than those among bus voltage phase angles. Therefore, when bus voltage phase data are unavailable, we can still achieve high accuracy of topology estimation using voltage magnitude. In Section~\ref{sec:num}, multiple simulation results show that using $\Delta |V|$ can provide accurate distribution grid topology estimation.} A detailed discussion of (\ref{eq:V_lasso}) is given in Appendix~\ref{sec:V_lasso}.

\subsection{Choice of the Regularization Parameter $\lambda$}
\label{sec:lambda_path}
The choice of $\lambda$ is critical in lasso and group lasso problems because it affects the number of non-zero coefficients in $\boldsymbol{\beta}^{(i)}$ and number of non-zero vector $\boldsymbol{\gamma}_k^{(i)}$. When $\lambda$ is small, the penalty term has no effect and the solution is close to the ordinary least squares (OLS) solution. When $\lambda$ is large, some coefficients of $\widehat{\boldsymbol{\beta}}^{(i)}$ or some vectors $\widehat{\boldsymbol{\gamma}}_k^{(i)}$ are zeros. A well-known criterion to choose the parameter $\lambda$ is Bayesian information criterion (BIC). For bus $i$, the BIC is defined as 

\begin{equation}
\label{eq:BIC_i} 
\text{BIC}_i(\lambda) = \frac{\text{RSS}_i(\lambda)}{\widetilde{T}\widehat{\sigma}^2} + \frac{\ln\widetilde{T}}{\widetilde{T}}\times \hat{k},	
\end{equation}
where $\hat{k}$ denotes the number of non-zero elements in $\widehat{\boldsymbol{\beta}}^{(i)}$ or the number of non-zero vector $\widehat{\boldsymbol{\gamma}}_k^{(i)}$, and $\widehat{\sigma}^2$ denotes the empirical variance of the residual \cite{zou2007degrees}. $\widetilde{T}$ is $2T$ for the problem in (\ref{eq:group_lasso}) or $T$ for the problem in (\ref{eq:V_lasso}). The residual sum of squares (RSS) is defined as
\[
\text{RSS}_i(\lambda) = \sum_{t=1}^T \left\|\mathbf{z}_i[t] - \sum_{\substack{k=2 \\ k \neq i}}^M \mathbf{x}_k[t]\widehat{\boldsymbol{\gamma}}^{(i)}_k\right\|_2^2.
\]
We select the $\lambda$ that minimizes $\text{BIC}_i(\lambda)$ as the optimal tuning parameter for bus $i$. The selection process requires to solve the problem in (\ref{eq:group_lasso}) or (\ref{eq:V_lasso}) multiple times. Thankfully, many lasso solvers, such as the least angle regression (LAR) \cite{efron2004least,yuan2006model}, can solve the lasso problem with multiple $\lambda$s at once. Therefore, this selection process can be completed without any additional computation. For (\ref{eq:joint_OR}) and (\ref{eq:joint_AND}), we can use the same approach to choose $\lambda$.

Fig.~\ref{fig:lambda_path} shows the path of BIC in each step of LAR algorithm. At each step, the LAR algorithm chooses a $\lambda$ and computes the corresponding coefficients $\widehat{\boldsymbol{\gamma}}^{(i)}$. Then, it decreases $\lambda$ and repeats the process above. Therefore, at Step 1, $\lambda$ has the largest value and all coefficients are zero. For the last step, all coefficients are non-zero. In Fig.~\ref{fig:lambda_path}, we can observe that the proposed scheme finds the sparse coefficient vector. Notice that in Fig.~\ref{fig:lambda_path}, we do not pick up $\lambda$ that yields the minimum BIC because the estimated coefficient vector has no sparsity, e.g., all elements in the coefficient estimate  are non-zero. Hence, we choose $\lambda$ that reduces BIC significantly with high sparsity.
\putFig{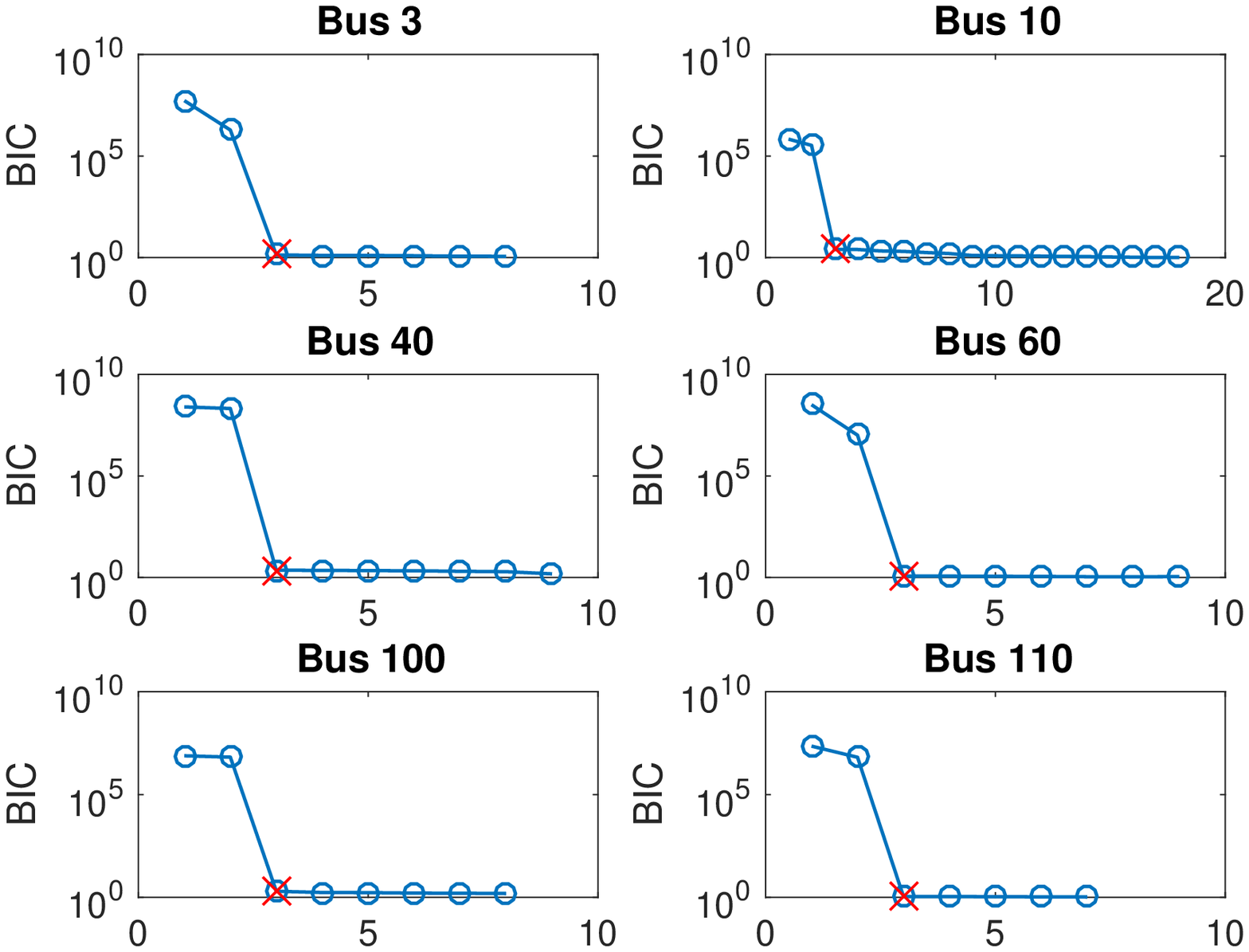}{BIC at each step for computing the bus connectivity  in IEEE 123-bus system. The circle represents the BIC at each step. The red crossing represents the corresponding BIC of the selected $\lambda$.}{\linewidth}

\section{Grid Topology Estimation via Group Lasso}
\label{sec:joint}
In the previous section, we have used group lasso to estimate the bus connectivity. However, a sub-network contains multiple unknown branches. In this section, we will extend the presented method from a bus to a network. 

The neighbor set $\mathcal{N}(i)$ can be found by solving the group lasso problem of bus $i$. \rev{Using the neighbor set of each bus, we can find the unknown branch between two buses. In \cite{meinshausen2006high}, two rules are proved to find the unknown edges in graphical models with a guarantee of false alarm rate: AND rule and OR rule. Specifically, if bus $i$ and $k$ are adjacent, bus $i$ is a neighbor of bus $k$ and vice versa.} In (\ref{eq:group_lasso}), we find the neighbors of a single bus. Therefore, an edge between bus $i$ and $k$, $e_{ik}$, is estimated twice since $k \in \mathcal{N}(i)$ and $i \in \mathcal{N}(k)$. A simple approach is combining them, i.e., $\widehat{e_{ik}}^\text{AND} = \widehat{\boldsymbol{\beta}}^{(i)}_k \wedge \widehat{\boldsymbol{\beta}}^{(k)}_i$, where $\wedge$ denotes the logical ``and''. We can apply this AND rule to every pair of buses within the unknown subnetwork and recover the topology at last.

Using the AND rule, an edge can only be recovered if both $\widehat{\boldsymbol{\beta}}^{(i)}_k$ and $\widehat{\boldsymbol{\beta}}^{(k)}_i$ are nonzero. If either of them has an error, this edge will be not included. Therefore, the AND rule has a high probability to miss the true edge. To minimize the number of missing edges, we propose the OR rule, $ \widehat{e}_{ik}^\text{OR} = \widehat{\boldsymbol{\beta}}^{(i)}_k \vee \widehat{\boldsymbol{\beta}}^{(k)}_i$, where $\vee$ is the logical ``or'' operator.

\rev{Although both AND and OR rules have been statistically proven to estimate unknown edges \cite{meinshausen2006high}, they are not been proven to satisfy the power system constraint that the estimated graph is a connected network.} Some buses may create an independent graph and are isolated from the main grid. To overcome this \rev{power system constraint}, we merge both rules together and propose the AND-OR rule. Unlike the AND or OR rule, this new rule has multiple steps. We firstly use the AND rule to estimate the edge set $\widehat{\mathcal{E}}_\text{AND}$. Then, we diagnose and adjust $\widehat{\mathcal{E}}_\text{AND}$ by physical law. In distribution grid, we assume that the differences between bus voltage magnitudes have a strong impact on the direction of power flow \cite{coffrin2014linear}. Therefore, each load bus is expected to have a neighbor bus with higher voltage magnitude on average because most load buses absorb powers. If none of the bus $i$'s neighbors has higher voltage magnitude, bus $i$ is possibly isolated from the main grid. We can find a new neighbor for bus $i$ by apply the modified OR rule, i.e.:
\rev{
\begin{equation}
\label{eq:AND_OR_rule}
\widehat{e}_{ik}^\text{AND-OR} = (\widehat{\E}(|V_k|) > \widehat{\E}(|V_i|)) \wedge (\widehat{\boldsymbol{\beta}}^{(i)}_k \vee \widehat{\boldsymbol{\beta}}^{(k)}_i),
\end{equation}
}
where $\widehat{\E}$ denotes the sample mean. The estimated edge set is $\widehat{\mathcal{E}}_\text{AND-OR} = \widehat{\mathcal{E}}_\text{AND} \cup \{\widehat{e}_{ik}^\text{AND-OR}\}$. Please note that we use $|V|$, not $\Delta |V|$, to enforce the AND-OR rule. The steps of AND-OR rule are summarized in Algorithm~\ref{alg:and-or}. We will show that the AND-OR rule provides more accurate and robust estimates than either the AND or OR rule in Section~\ref{sec:num}.
\begin{algorithm}[H]
 \caption{Distribution Grid Topology Estimation via the AND-OR Rule}
\begin{algorithmic}[1]
\REQUIRE Topology estimate of the AND rule $\widehat{\mathcal{E}}_\text{AND}$
\IF {$\exists k \in \widehat{\mathcal{N}}(i)$ satisfies $ \widehat{\E}(|V_k|) > \widehat{\E}(|V_i|)$}
\STATE {Find the new edge $\widehat{e}_{ik}^\text{AND-OR}$ using (\ref{eq:AND_OR_rule})}
\STATE {$\widehat{\mathcal{E}}_\text{AND-OR} = \widehat{\mathcal{E}}_\text{AND} \cup \{\widehat{e}_{ik}^\text{AND-OR}\}$}
\ELSE
\STATE {$\widehat{\mathcal{E}}_\text{AND-OR} = \widehat{\mathcal{E}}_\text{AND}$} 
\ENDIF
\end{algorithmic}
\label{alg:and-or}
\end{algorithm}

%
%
%

\subsection{Grid Topology Estimation via Group Lasso}
While the AND-OR rule is robust in reconstructing distribution grid topology, this process requires finding the bus connectivity of each bus at first. This leads to high computational time for large systems, due to solving multiple optimization problems. Also, it has not utilized the interactions between bus measurements. To improve the algorithm, we will formulate the grid topology estimation as a single optimization problem via group lasso.

Let's start with a simple case. We assume only voltage magnitude data $\Delta |V|$ are available. In a fully connected $3$-bus system, 
we can express the voltage relationships using the following linear system,
\begin{eqnarray}
	\Delta |\mathbf{V}| &=& \mathbf{P}\widetilde{\boldsymbol{\beta}},\label{eq:joint_linear_sys} \\
	\begin{bmatrix}
		\Delta | V_1| \\
		\Delta | V_2| \\
		\Delta | V_3|
	\end{bmatrix}
	&=&
	\begin{bmatrix}
		\Delta | V_2| & 0 & 0\\
		0 & \Delta | V_1| & 0 \\
		\Delta | V_3| & 0 & 0\\
		0 & 0 & \Delta | V_1| \\
		0 & \Delta | V_3| & 0 \\
		0 & 0 & \Delta | V_2|      
	\end{bmatrix}^T
	\begin{bmatrix}
		\widetilde{\beta}^{(1)}_2 \\
		\widetilde{\beta}^{(2)}_1 \\
		\widetilde{\beta}^{(1)}_3 \\
		\widetilde{\beta}^{(3)}_1 \\
		\widetilde{\beta}^{(2)}_3 \\
		\widetilde{\beta}^{(3)}_2
	\end{bmatrix}. \nonumber
\end{eqnarray}
Estimating the distribution grid topology by the AND or OR rule is equivalent to solving the linear system in (\ref{eq:joint_linear_sys}) with different penalties. Specifically, for an $M$-bus system, the following optimization problem is equivalent to the OR rule:
\begin{equation}
\label{eq:joint_OR}
	\widehat{\boldsymbol{\beta}}^\text{OR} = \arg\min_{\widetilde{\boldsymbol{\beta}}}\sum_{t=1}^T\left\|\Delta | \mathbf{v}[t]| - \mathbf{P}[t]\widetilde{\boldsymbol{\beta}}\right\|_2^2 + \lambda\|\widetilde{\boldsymbol{\beta}}\|_1,
\end{equation}
where $\Delta | \mathbf{v}[t]| \in \reals^{(M-1)\times 1}$, $\mathbf{P}[t] \in \reals^{(M-1)\times (M-1)(M-2)}$, and $\widetilde{\boldsymbol{\beta}} \in \reals^{(M-1)(M-2)\times 1}$. With the $L_1$ norm penalty, any element in $\widetilde{\boldsymbol{\beta}}$ can be zero. Therefore, (\ref{eq:joint_OR}) is equivalent to the OR rule. When $M$ is large, the dimension of $\mathbf{P}[t]$ is high. However, since $\mathbf{P}[t]$ is sparse and only contains $M$ unique values, we can store and process it efficiently. In Section~\ref{sec:compute}, we will further simplify $\mathbf{P}[t]$.

For the AND rule, $\widetilde{\beta}^{(i)}_k$ and $\widetilde{\beta}^{(k)}_i$ are either zero or non-zero simultaneously. Therefore, we can use the group lasso to solve problem in (\ref{eq:joint_linear_sys}). In details, we can solve the following optimization problem:
\begin{equation}
\label{eq:joint_AND}
	\widehat{\boldsymbol{\beta}}^\text{AND} = \arg\min_{\widetilde{\boldsymbol{\beta}}}\sum_{t=1}^T\left\|\Delta | \mathbf{v}[t]| - \mathbf{P}[t]\widetilde{\boldsymbol{\beta}}\right\|^2 + \lambda \sum_{\substack{i,k=2 \\ k \neq i}}^M \|(\widetilde{\beta}^{(i)}_k,\widetilde{\beta}^{(k)}_i)\|_2.
\end{equation}
In (\ref{eq:joint_AND}), we group all bus pairs into a penalty term to decide the pairwise connectivity. Thus, (\ref{eq:joint_AND}) is equivalent to the AND rule. To apply the AND-OR rule, we can follow the same step shown in Algorithm.~\ref{alg:and-or}.

\section{Simulation and Results}\label{sec:num}
We firstly use IEEE $8$-bus and $123$-bus networks \cite{teng2002modified,kersting2001radial}, with additional branches to create mesh structures, to validate the performance of our lasso-based approach on mesh networks. Fig.~\ref{fig:new_8bus} illustrates the modified mesh $8$-bus system. Also, we justify our methods on six European representative distribution systems with different topologies, which include MV and LV distribution grids in urban (\textit{LV\_urban}, \textit{MV\_urban}, \textit{MV\_two\_substations}, \textit{Urban}), suburban (\textit{LV\_suburban}) and rural (\textit{MV\_rural}) areas\cite{pretticodistribution}. To explore the impact of loops, we add several branches in LV and MV grids and generate \textit{LV\_suburban\_mesh} and \textit{MV\_urban\_mesh} systems. \textit{Urban} system is a large-scale distribution grid that includes MV and LV networks. It has about 13,000 customers, 126 MV/LV substations, and 3237 branches. Most urban and suburban branches in these networks are underground. The European system topology and details are available in Appendix~\ref{sec:eu_network}. In each network, the feeder or substation is selected as the slack bus.


The smart meter hourly load readings from PG\&E are used in all simulations. Since PG\&E data set does not have the reactive power, we emulate $q_i[t]$ according to a random lagging power factor $pf_i[t]$, e.g., $pf_i[t] \sim \Unif(0.85,0.95)$. For the load profile in MV grid, we aggregate the load profiles of 10 to 300 residents, depending on the load capacity. The hourly voltage measurements $v_i[t]$ are obtained by MATPOWER\cite{Zimmerman10} and $N = 8760$ measurements are computed at each bus.
 

\subsection{Estimation Error of Bus Connectivity}
For bus $i$, we define the connectivity error as
\[
\text{Error}(i) = \underbrace{\sum_{k \in \mathcal{N}(i)} \indic{k \notin \widehat{\mathcal{N}}(i)}}_{\text{false estimation}} + \underbrace{\sum_{k \in \widehat{\mathcal{N}}(i)} \indic{k \notin \mathcal{N}(i)}}_{\text{missing}},
\]
where $\widehat{\mathcal{N}}(i)$ denotes the neighbor set estimate using (\ref{eq:group_lasso}) or (\ref{eq:V_lasso}) and $\indic{}$ denotes the indicator function. The first part represents the number of missing neighbors and the second part represents the number of incorrect neighbors.

\begin{table}[h!]
\caption{Total Branch Error of Lasso (\ref{eq:Lasso}) and Group Lasso (\ref{eq:group_lasso}) using $\Delta V$}
\centering
\begin{tabular}{|c|c|c|}
\hline
	System & Lasso & Group Lasso \\
	\hline
	123-bus & 10 & 0 \\
	123-bus with loops & 10 & 1 \\
	123-bus with PV & 6 & 0 \\
	123-bus with loops \& PV & 6 & 0 \\
\hline
\end{tabular}
\label{tab:complex_lasso}
\end{table}

Table~\ref{tab:complex_lasso} shows the total branch estimation error of lasso (\ref{eq:Lasso}) and group lasso (\ref{eq:group_lasso}) using $\Delta V$. Without grouping the real and imaginary parts, the lasso method has worse performance because of the inconsistency between real and imaginary parts of the complex estimate. By adding grouping constraint, our method achieves nearly perfect results.

Our simulations show that for 8-bus networks, with or without loops, our algorithm presented in Section~\ref{sec:main} achieves zero error using $\Delta |V|$. Fig.~\ref{fig:nodeMissing123} shows the error at each bus for 123-bus networks, with or without loops. We can observe that most buses have zero error by using $\Delta |V|$. While identifying all connectivities in a $123$-bus system is excessive in practice, our method can find all connectivities successfully except two or three buses. For the European representation network, we observe the similar performance as IEEE systems.
\putFig{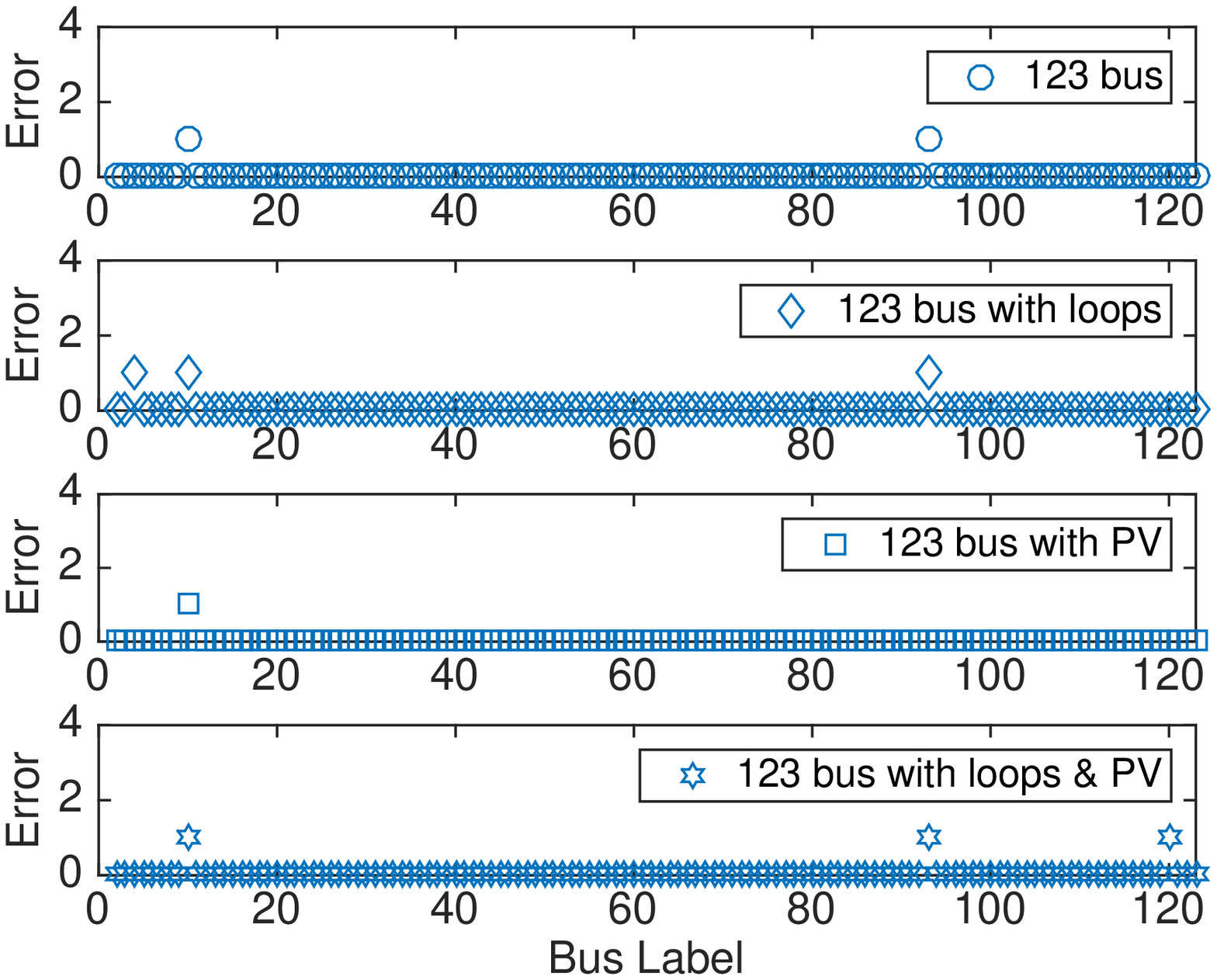}{Errors of bus connectivity reconstruction for different networks using $\Delta | V|$ only.}{\linewidth}

For the bus connectivity estimation with integrated DERs, our simulations show that the proposed approach finds the buses' neighbors without any error in 8-bus system. In 123-bus networks, as shown in Fig.~\ref{fig:nodeMissing123}, our approach has nearly perfect performance.

\subsection{Network Topology Reconstruction Error Rate}
In this section, we discuss the performance on grid topology reconstruction. We use the error rate (ER) as the performance evaluation metric, which is defined as
\begin{eqnarray*}
\text{ER} &=& \frac{1}{|\mathcal{E}|}\left(\underbrace{\sum_{e_{ij} \in \widehat{\mathcal{E}}} \indic{e_{ij} \notin \mathcal{E}}}_{\text{false estimation}} + \underbrace{\sum_{e_{ij} \in \mathcal{E}} \indic{e_{ij} \notin \widehat{\mathcal{E}}}}_{\text{missing}}\right)\times 100\% 
\end{eqnarray*}
where $\widehat{\mathcal{E}}$ denotes the edge set estimates, $|\mathcal{E}|$ is the size of $\mathcal{E}$, and $\indic{.}$ is the indicator function. The first and second terms represent the number of falsely estimated branches and the number of missing branches, respectively.

In Table~\ref{tab:mismatch_group}, we summarize the error rates of 123-bus systems and six EU representative distribution grids with different topology configurations and decision rules. When the system is integrated with PMUs, the group lasso method reconstructs most MV and LV systems with zero error. When only voltage magnitude is available, the performances of OR and AND rules are degraded. But the AND-OR rule still achieves nearly perfect performance in most test cases. For \textit{Urban} system, which contains both MV and LV grids, the majority of the error is due to the missing edges. Only four branches are incorrectly estimated. These missing edges are ones that connect MV/LV transformers. For many utilities, the location information of these transformers is known. With the locational information, the error rate is reduced to $0.6\%$. A similar improvement can be obtained for \textit{Urban} with all switches closed. To validate our algorithm on large-scale distribution grids without transformers, we create an artificial grid (\textit{LV\_large}) by combing 31 \textit{LV\_suburban\_mesh} grids. From Table~\ref{tab:mismatch_group}, we observe that the error does not scale by the size of grid.

In some distribution grids, the voltage of slack bus may not be constant. As discussed in Section~\ref{sec:formulation}, we can model the slack bus voltage as a random variable ($\Delta V_1$) and include it in (\ref{eq:joint}). To validate this case, we perform a simulation on a LV grid in \textit{Urban} system. In this simulation, the feeder is selected as the slack bus and its voltage is determined by the upper MV grid. Hence, the voltage measurements of slack bus are not constant. By utilizing the proposed algorithm, the error rate does not change when the slack bus voltage is not constant.

\begin{table}[h!]
\caption{Network Topology Reconstruction Error Rate without DERs}
\centering
\begin{tabular}{|c||c|c|c|c|}
\hline
System & Total  & AND & OR & AND-OR \\
&Branch & $\Delta |V|$ & $\Delta |V|$ & $\Delta |V|$ \\
\hline
$8$-bus & 7 & 0\% & 14.29\% & 0\% \\
\hline
$8$-bus & 10  & 20\% & 10\% & 0\% \\
3 loops &&&& \\
\hline
$123$-bus & 122  & 4.07\% & 2.44\% & 0\% \\
\hline
$123$-bus  & 124  & 4.07\% & 1.63\% & 0\% \\
2 loops &&&& \\
\hline
\textit{LV\_urban} & 13   & 0\% & 0\% & 0\% \\
\hline
\textit{LV\_suburban} & 114  & 4.42\% & 0.88\% & 0\%\\
\hline
\textit{LV\_suburban\_mesh} & 129 & 2.33\% & 8.53\% & 0.78\%\\
15 loops &&&& \\
\hline
\textit{MV\_urban} & 34  &  0\% & 5.88\% & 0\% \\
\hline
\textit{MV\_urban} & 35 & 0\% & 0\% & 0\%\\
switch 34-35 &&&& \\ 
1 loop &&&& \\
\hline
\textit{MV\_urban} & 35  & 2.86\% & 5.71\% & 2.86\% \\
switch 23-35 &&&& \\ 
1 loop &&&& \\
\hline
\textit{MV\_urban} & 35  & 2.86\% & 5.71\% & 0\% \\
switch 13-35 &&&& \\ 
1 loop &&&& \\
\hline
\textit{MV\_urban} & 37  & 0\% & 0\% & 0\% \\
3 switches &&&& \\ 
3 loops &&&& \\
\hline
\textit{MV\_urban\_mesh} & 44  &  0\% & 4.55\% & 0\% \\
10 loops &&&& \\
\hline
\textit{MV\_two\_stations} & 46 & 4.35\% & 8.7\% & 0\% \\
\hline
\textit{MV\_two\_stations} & 47  & 4.25\% & 12.77\% & 0\% \\
switch 14-37 &&&& \\
1 loop &&&& \\
\hline
\textit{MV\_two\_stations} & 47  & 0\% & 0\% & 0\% \\
switch 24-48 &&&& \\
1 loop &&&& \\
\hline
\textit{MV\_two\_stations} & 48  & 0\% & 6.25\% & 0\% \\
2 switches &&&& \\
2 loops &&&& \\
\hline
\textit{MV\_rural} & 116 & 0\% & 11.30\% & 0\% \\
\hline
\textit{MV\_rural} & 119 & 0\% & 11.02\% & 0\% \\
3 switches &&&& \\
3 loops &&&& \\
\hline
\textit{Urban} & 3237 & 19.62\% & 10.72\% & 6.98\% \\
\hline
\textit{Urban} & 3242 & 19.56\% & 10.61\% & 6.97\% \\
all switches &&&& \\
5 loops &&&& \\
\hline
\textit{LV\_large} & 4030 & 0.99\% & 5.98\% & 0.12\%\\
465 loops &&&& \\
\hline
\end{tabular}
\label{tab:mismatch_group}	
\end{table}

\putFig{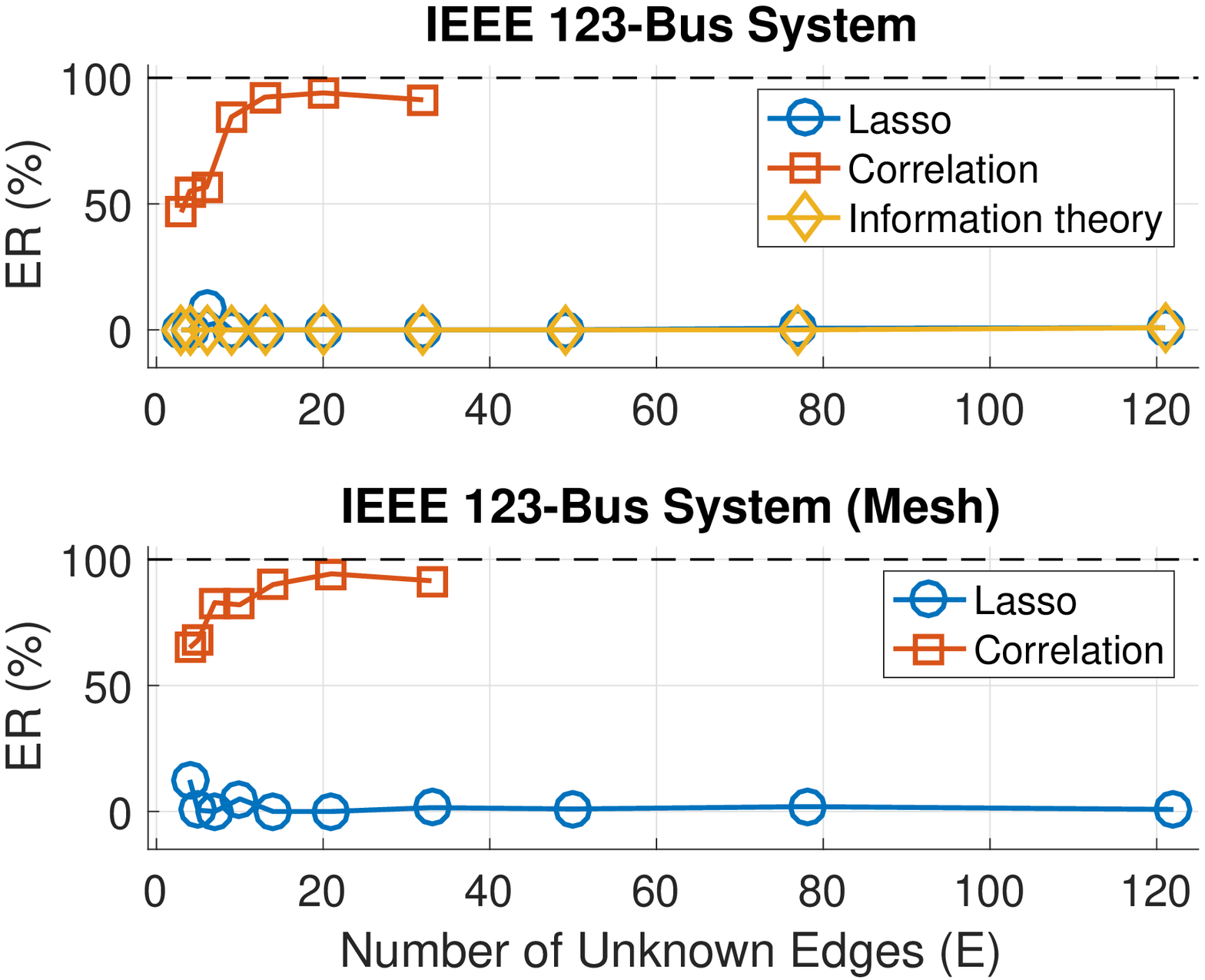}{Comparison of ER amongst three methods using $\Delta |V|$ in IEEE 123-bus systems with radial and mesh structures.}{0.8\linewidth}

The performance comparison amongst our lasso-based algorithm, a correlation-based algorithm \cite{bolognani2013identification}, and an information theory-based algorithm \cite{liao2015distribution} is illustrated in Fig.~\ref{fig:lasso_compare2_all}. \rmtxt{The optimization problem in \cite{bolognani2013identification} is solved by CVX, a package for specifying and solving convex programs \cite{cvx}. The $x$-coordinate represents the number of edges that are needed to be identified. The $y$-coordinate represents the average error rate over 100 iterations.} The error rates are averaged over $100$ iterations. The proposed method consistently recovers the topology with nearly $0\%$ error rate. This result is comparable with  \cite{liao2015distribution}, while the detection ability of \cite{bolognani2013identification} drops. The approach in \cite{liao2015distribution} is excluded in the mesh network comparison because it can only be applied to radial networks.


\subsection{Networks with DER Integration}
The penetration of DERs has grown significantly during last decade and will keep increasing in the future. To evaluate the proposed algorithm with integrated DERs, we install several rooftop photovoltaic (PV) systems in the distribution networks. The profile of hourly power generation is obtained from NREL PVWatts Calculator, an online simulator that estimates the PV power generation based on weather history of PG\&E service zone and the physical parameters of a $5$kW PV panel in LV grids or a $20$kW PV panel in MV grids\cite{dobos2014pvwatts}. The power factor is fixed as $0.90$ lagging, which satisfies the regulation of many U.S. utilities \cite{ellis2012review} and IEEE standard \cite{ieee2014guide}.


\begin{table}[h]
\caption{Network Topology Reconstruction Error Rate with Rooftop PV System using $\Delta |V|$}
\centering
	\begin{tabular}{|c||c|c|c|c|c|}
	\hline
	System & Total PV & AND & OR & AND-OR \\
	\hline
	8-bus & 8 & 14.29\% & 14.29\% & 0\% \\
	\hline
	8-bus & 8 & 10.00\% & 10.00\% & 0\% \\
	3 loops &&&& \\
	\hline
	123-bus & 12 & 2.44\% & 0.81\% & 0\% \\
	\hline
	123-bus & 12 & 4.07\% & 0\% & 0\% \\
	2 loops &&&& \\
	\hline
	\textit{LV\_suburban} & 10 & 0.88\% & 8.85\% & 0\% \\
	\hline
	\textit{LV\_suburban} & 20 & 2.65\% & 12.39\% & 0.88\% \\
	\hline
	\textit{LV\_suburban} & 33 & 4.42\% & 7.96\% & 0.88\% \\
	\hline
	\textit{MV\_urban} & 7 & 2.78\% & 2.78\% & 0\% \\
	\hline
	\textit{MV\_urban} & 7 & 0\% & 2.86\% & 0\%\\
	switch 34-35 &&&& \\ 
	1 loop &&&& \\
	\hline
	\textit{MV\_urban} & 7  & 0\% & 2.70\% & 0\% \\
	3 switches &&&& \\ 
	3 loops &&&& \\
	\hline
	\textit{MV\_two\_stations} & 10 & 4.35\% & 4.35\% & 0\% \\
	\hline
	\textit{MV\_two\_stations} & 10 & 0\% & 2.08\% & 0\% \\
	2 switches &&&& \\
	2 loops &&&& \\
	\hline
	\textit{MV\_rural} & 20 & 5.17\% & 12.07\% & 0.86\% \\
	\hline
	\textit{MV\_rural} & 20 & 11.86\% & 15.25\% & 2.52\% \\
	3 switches &&&& \\
	3 loops &&&& \\
	\hline
	\textit{Urban} & 300 & 19.96\% & 10.10\% & 9.02\% \\
	\hline
	\textit{LV\_large} & 300 & 1.41\% & 3.05\% & 0.42\%\\
	465 loops &&&& \\
	\hline
	\end{tabular}
\label{tab:mismatch_renewable}
\end{table}

The error rates of grid topology reconstruction with the rooftop PVs integration are presented in Table~\ref{tab:mismatch_renewable}. OR rule and AND rule have performance degradation. The error rate of the AND-OR rule is still the lowest one and most network topologies can be recovered perfectly. Additionally, we can observe that different levels of PV penetration do not have a significant impact on the algorithm performance. For \textit{Urban}, with the prior knowledge of transformer locations, the AND-OR rule misses $22$ branches amongst $3237$ branches and has no false estimation error. The error rate reduces to $0.68\%$. For \textit{LV\_large} system, the error rate of AND-OR rule is similar to the case without DERs in Table~\ref{tab:mismatch_group}.

\subsection{Computational Complexity}
\label{sec:compute}
The least angle regression (LAR) has a computational complexity of $\mathcal{O}(M^3+TM^2)$, where $T$ denotes the number of observation and $M$ denotes the grid size. Since the complexity is dominated by $M$, finding the bus connectivity takes a long time in a large-scale system. Fortunately, for a particular bus in the distribution grid, the number of neighbors is relatively small compared with the grid size \cite{pagani2011towards}. Therefore, to estimate the connectivity of a particular bus, we select $K$ buses from an $M$-bus system and then apply (\ref{eq:V_lasso}) only to these $K$ buses. The complexity for bus connectivity estimation reduces to $\mathcal{O}(K^3+TK^2)$. For the entire system, the complexity is $\mathcal{O}(M(K^3+TK^2))$, which is linear in term of the system size $M$. Also, the dimension of $P[t]$ in (\ref{eq:joint_OR}) and (\ref{eq:joint_AND}) becomes $(K-1)\times(K-1)(K-2)$, which is smaller than the full model and independent of the system size. For the systems we analyze in this paper, the average number of neighbor per bus is $2$, and the maximum is $10$. We choose $K=\sqrt{M}$ to ensure that the true neighbors are contained within these $K$ buses. \rmtxt{How to find $K$ most relative buses are discussed in \cite{liao2017lasso_su}.} 

In this paper, we use the mutual information as the measure to choose the $K$ most relative buses. Specifically, in \cite{liao2015distribution}, the authors prove that if two buses are connected, their mutual information is higher than the pair of buses that are not connected. Therefore, to solve the lasso problem efficiently, firstly, we compute the pairwise mutual information. Secondly, for bus $i$, we find the top $K$ buses that have the largest mutual information with bus $i$. Thirdly, we apply group lasso to these $K$ buses and find the neighbors of bus $i$.
\putFig{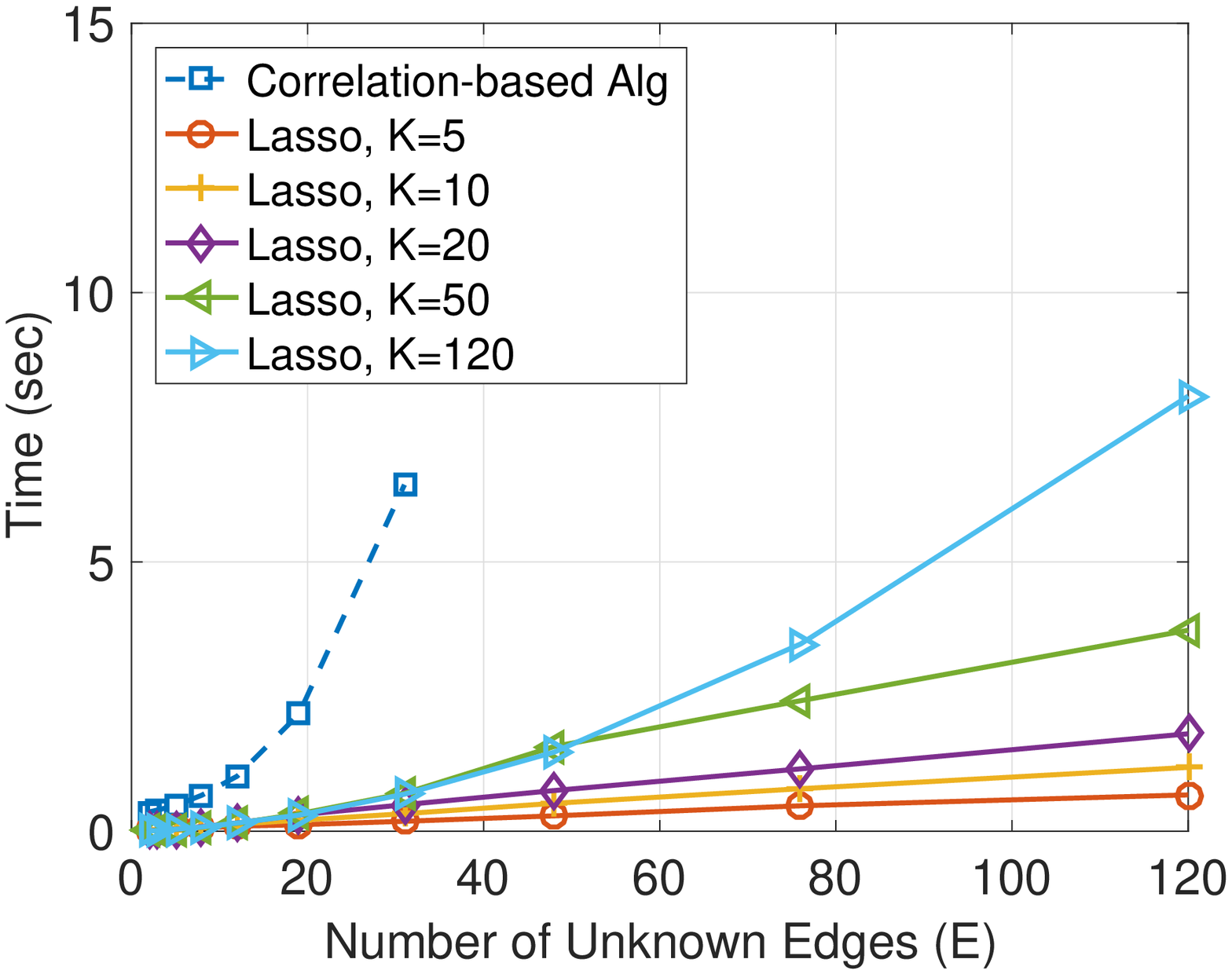}{Total computational time of the correlation-based algorithm and our method with different bus selections. We use $\Delta |V|$ and IEEE $123$-bus system with mesh structures.}{0.8\linewidth}

The average computational time of our lasso-based method and the correlation-based algorithm is summarized in Fig.~\ref{fig:timeComp}. We use LAR method to solve the problem in (\ref{eq:V_lasso}) and apply AND-OR rule to find the topology. Fig.~\ref{fig:timeComp} shows our method is consistently faster than \cite{bolognani2013identification}. Also, by selecting $K$ most relative buses, the proposed algorithm is faster by a factor of 12 and achieves the same accuracy. For the $8$-bus system with mesh structures in Fig.~\ref{fig:new_8bus}, our method only uses $0.3$ seconds to recover the topology. In \cite{weng2016distributed}, the authors extend the information theory-based algorithm, which is designed for radial network, to mesh grids. The extended algorithm requires over $100$ seconds to estimate the topology of mesh $8$-bus network. Hence, our approach is faster than other ones in mesh systems. For \textit{Urban} system with $K=30$, the average computational time is less than $270$ seconds, making it useful for semi-real time applications.

\subsection{Sensitivity Analysis}
In this subsection, we will discuss how data accuracy, data length, load pattern, and data resolution affect the algorithm performance.
\subsubsection{Sensitivity to Data Accuracy}
Our method relies on the smart meter measurements. Hence, it is important to know whether the existing meters' accuracy is sufficient for topology estimation. In the U.S., ANSI C12.20 standard (Class 0.5) requires the smart meters to have an error less than $\pm 0.5\%$ \cite{ansc12,zheng2013smart}. Table~\ref{tab:noise} shows the average error rate with different noise levels over 20 iterations. The AND-OR rule outperforms other two rules in all levels of noise. For the 123-bus system with loops and \textit{MV\_urban}, the AND-OR rule consistently reconstructs the entire network without any error. The accuracy of \textit{LV\_suburban} is reduced when the noise level is high.

\begin{table}[h!]
	\centering
	\caption{Error Rate With Different Noise Levels using $\Delta|V|$}
	\begin{tabular}{|c||c|c|c|c|c|c|}
	\hline
	& \multicolumn{2}{c|}{AND} & \multicolumn{2}{c|}{OR} & \multicolumn{2}{c|}{AND-OR} \\
	\hline
	Noise Level & 0.1\% & 0.5\% & 0.1\% & 0.5\% & 0.1\% & 0.5\% \\
	\hline
	123-bus & 8.94\% & 10.28\% & 2.44\% & 2.44\% & 0\% & 0\% \\
	\hline
	\textit{LV\_suburban} & 7.96\% & 10.84\% & 4.42\% & 8.36\% & 0\% & 4.2\% \\
	\hline
	\textit{MV\_urban} & 5.41\% & 5.41\% & 0\% & 0\% & 0\% & 0\% \\
	all switches  &&&&&& \\
	closed &&&&&& \\
	\hline
	\end{tabular}
	\label{tab:noise}
\end{table}

In some countries, the utilities pre-process the voltage measurements, e.g., round up the float data to integers. These types of data processing can create identical measurements, e.g., $110$ volts, for a majority of time, making our algorithm relative poor due to the loss of a statistical relationship. However, our method can give recommendations to what data resolution is needed to utilize smart meter beyond billing purpose only.

In distribution grids, some switches may change statuses to protect circuits temperately. If we include the measurements collected during the temperate switch changes, our algorithm may have a decrease in accuracy. To overcome this issue, we can apply the method discussed in \cite{liao2016urban} to identify and remove these measurements. Then, we use the rest data to estimate grid topology. Also, as shown previously, our algorithm can estimate a large-scale grid in a few minutes. Thus, we can apply this algorithm to data sets acquired at consecutive time slots and check if the results are identical.

\subsubsection{Sensitivity to Data Length}
To understand the impact of data set size, we validate the proposed algorithm by using measurements from 2 up to 360 days. Fig.~\ref{fig:data_duration} shows the error rates of different networks with various data lengths. For the 123-bus loopy network, we can see that with around 100 days' measurements ($100\times24 = 2400$ data points), the AND-OR rule achieves zero error. For \textit{LV\_suburban\_mesh} and \textit{MV\_urban\_mesh}, only ten days' data are required to achieve perfect estimation. \rmtxt{In \cite{liao2017lasso_su}, we show that our algorithm requires less time to achieve good performance with a higher sampling frequency.}
\putFig{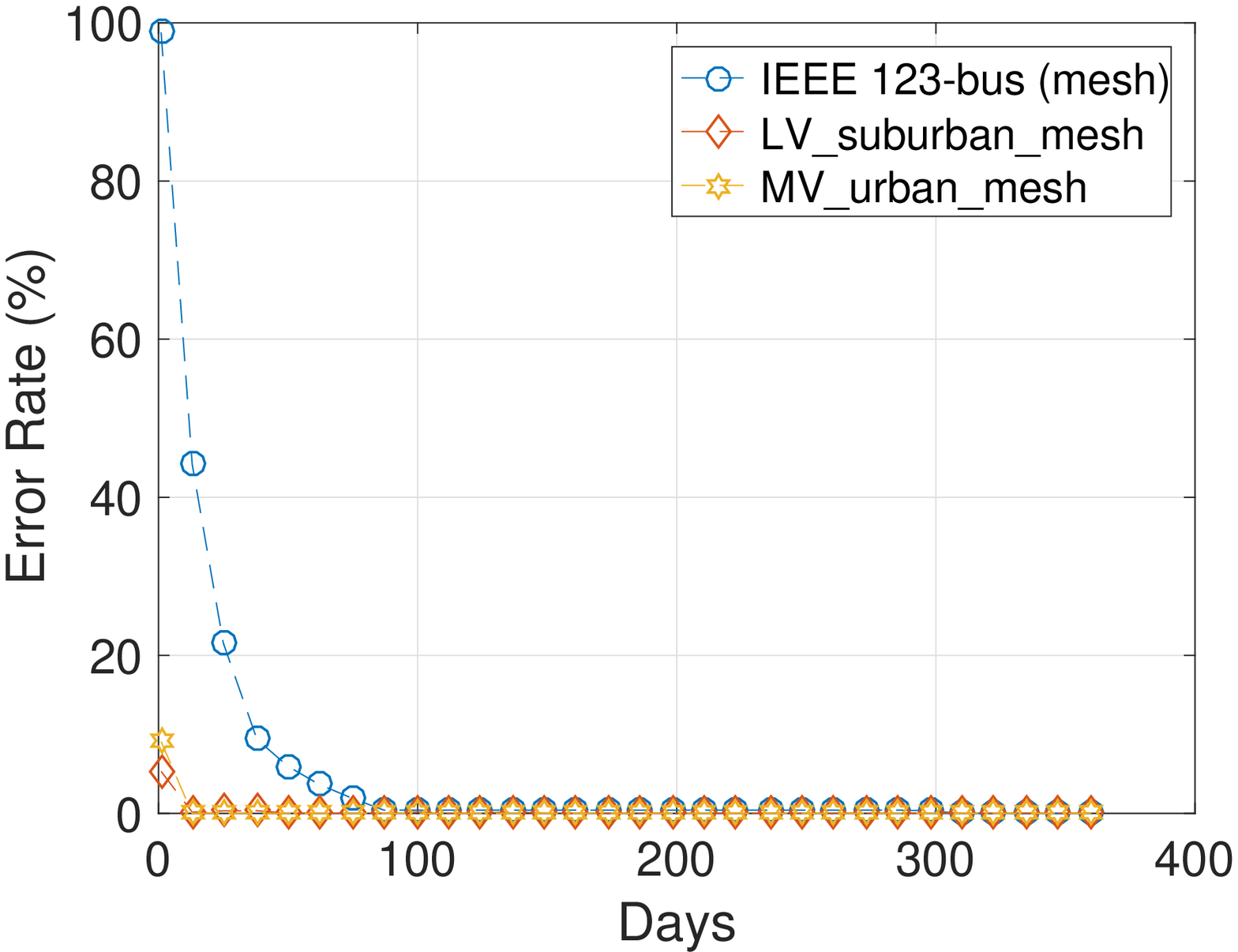}{Error rates of the AND-OR rule with different data lengths using $\Delta |V|$.}{0.8\linewidth}

\subsubsection{Sensitivity to Load Pattern}
To understand our algorithm's sensitivity to load pattern, we validate the proposed algorithm on the ``ADRES-Concept'' Project load profile \cite{Einfalt11,VUT16}. This data set contains real and reactive powers profile of 30 houses in Upper-Austria. The data were sampled every second over 7 days in summer and 7 days in winter.

In Fig.~\ref{fig:EU_winter_summer}, we compare the error rates using winter and summer load profiles individually. The voltage profiles are obtained by using IEEE 123-bus test case and assuming that the connectivity between bus 78 and bus 102 is unknown. From Fig.~\ref{fig:EU_winter_summer}, we observe that the data coming from different seasons do not impact the algorithm performance. Also, our algorithm converges faster to zero error.
\putFig{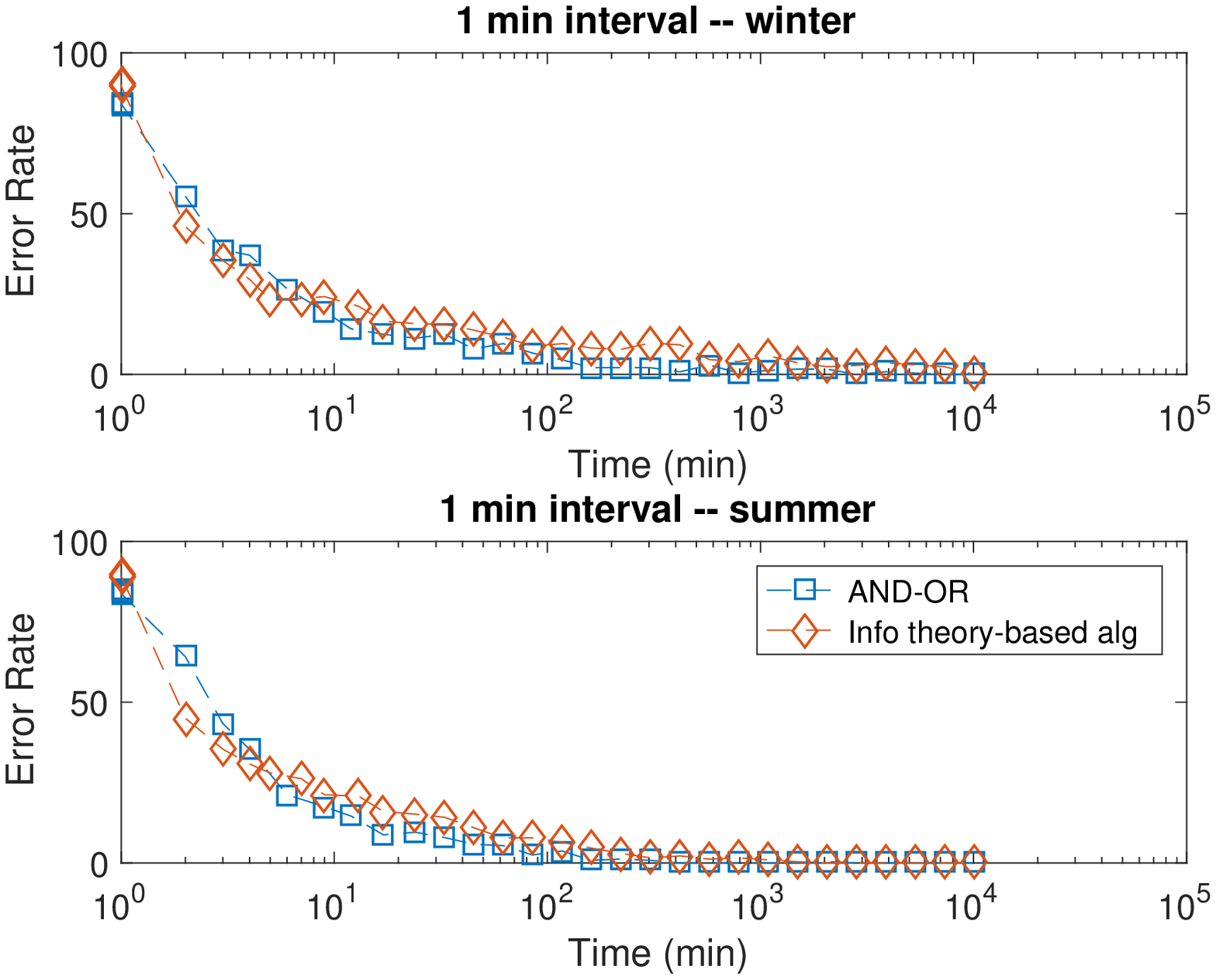}{Comparison of error rate with data from different seasons using $\Delta|V|$.}{\linewidth}

\subsubsection{Sensitivity to Data Resolution}
Fig.~\ref{fig:EU_time} illusrates the performance of AND-OR rule under different sampling frequencies. When the sampling period is 1 minute, we need about 2 hours' voltage profile to have perfect estimation of the entire system. The frequency of distribution grid reconfiguration is range from hours to weeks \cite{jabr2014minimum,dorostkar2016value}. Therefore, the proposed method is suitable for the existing system and real-time operation. If the sampling period is 30 minutes, our algorithm needs about 50 hours' measurements. This data requirement is still less than the current network reconfiguration frequency. Also, Fig.~\ref{fig:EU_time} shows that the estimation time can be reduced by using high sampling frequency.
\putFig{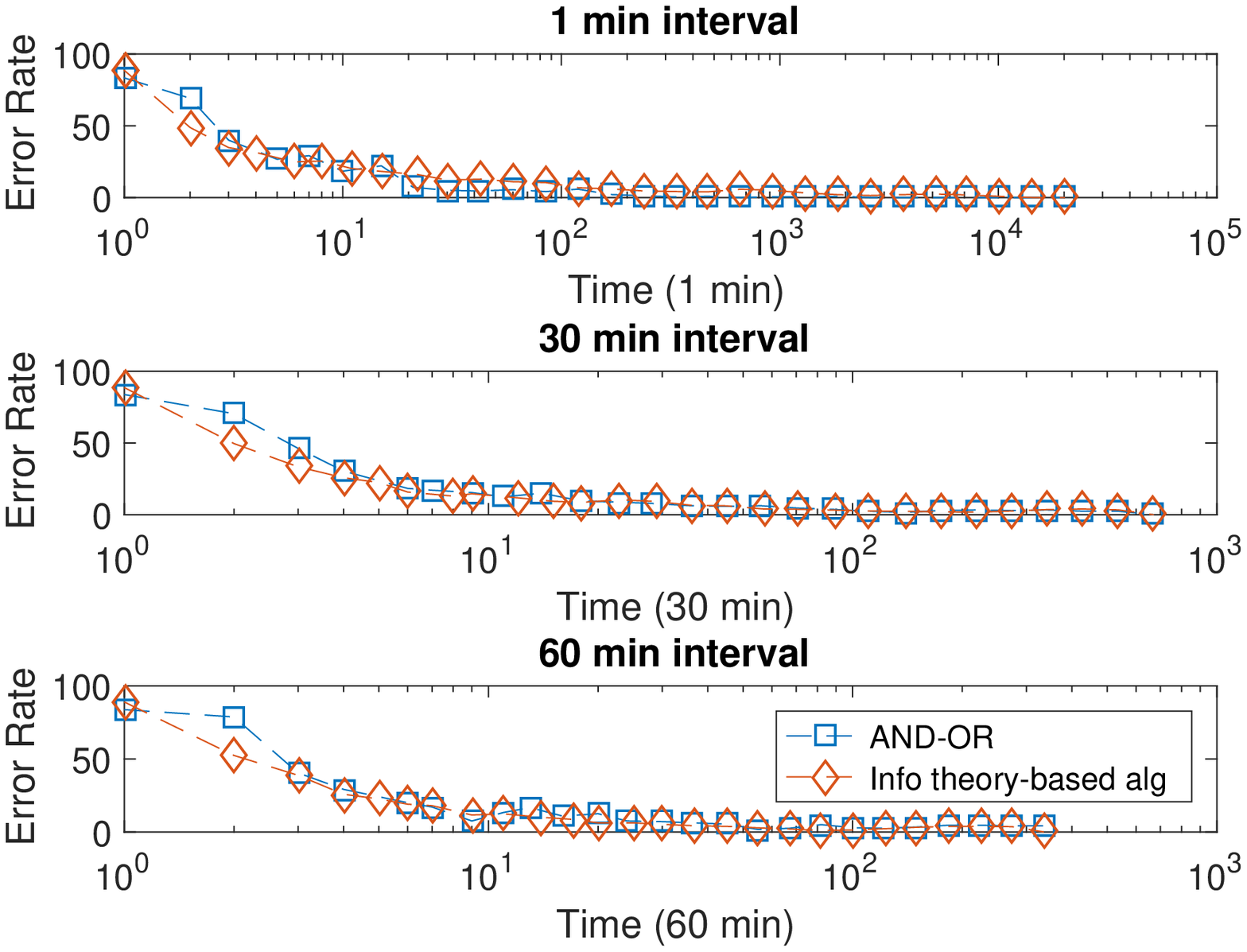}{Comparison of error rate with different data resolutions using $\Delta|V|$.}{\linewidth}

\section{Conclusion}\label{sec:con}
A data-driven algorithm of bus connectivity and grid topology estimation is presented in MV and LV distribution grids. Comparing with past studies, our method does not need the admittance matrices or switch location information. Only the smart meter data (voltage magnitude profile) are utilized for topology estimation. Also, unlike many past studies, our method can estimate not only radial systems but also mesh networks. We prove that, representing a distribution grid as a graphical model, the grid topology can be efficiently recovered by group lasso. We validate the proposed algorithm on eight MV and LV distribution networks with 21 network configurations using real data from PG\&E and NREL. With or without DER penetration, our algorithm estimates the topology of a large-scale MV/LV grid with over 95\% accuracy in a short period of computational time. Finally, we analyze the algorithm performance under different noise levels, data resolutions, data duration, and load patterns. The results indicate that our method can provide robust estimation in various scenarios and outperform other existing methods. \rmtxt{More details on algorithm performance are presented in the supplementary materials \cite{liao2017lasso_su}.}

\section{Acknowledgement}
The authors would like to acknowledge Dr. Giuseppe Prettico from European Commission Joint Research Centre for sharing the European Representative Distribution Networks. We would also like to thank Vienna University of Technology - Institute of Energy Systems and Electrical Drives for providing the ADRES-Concept data set. The first author would like to thank Dr. Junjie Qin from Stanford University for discussions and Stanford Leavell Fellowship for the financial support. Part of this project is supported by State Grid Corporation technology project (SGRIJSKJ(2016)800).

\appendix
\subsection{Proof of Lemma~\ref{lemma: linear_indept}}
\label{sec:lemma1_proof}

\rev{
\begin{IEEEproof}
We will prove Lemma~\ref{lemma: linear_indept} using a counterexample. Assuming $a_1 \neq 0$ and $b_1 \neq 0$, given $Y = y$, we have the following equation
\begin{eqnarray*}
	a_1 X_1 + b_1 X_2 &=& y, \\
	X_1 &=& (y- b_1 X_2)/a_1.
\end{eqnarray*}
Since $a_1$ and $b_1$ are non-zeros, $X_1$ always depends on $X_2$. Therefore, to have $X_1$ and $X_2$ conditionally independent, at least one coefficient needs to be zero, e.g., $a_1 = 0$, $b_1 = 0$, or $a_1 = b_1 = 0$.
\end{IEEEproof}
}

\subsection{Proof of Lemma~\ref{lemma:linear_indept2}}
\label{sec:lemma2_proof}
\rev{
\begin{IEEEproof}
	Assuming $c_1 = 0$, (\ref{eq:lemma2_1}) becomes $d_1 Z= X_1$. Given $Z = z$, $X_1 = d_1 z$ is a constant. Therefore, $X_1$ and $X_2$ are conditionally independent. When $c_2 = 0$ and $Z=z$, $X_2$ becomes a constant and therefore, $X_1$ and $X_2$ are conditionally independent.
	
	When $c_1 = 0$ and $c_2 = 0$, both $X_1$ and $X_2$ become constants. Therefore, they are not random variables and we cannot determine statistical dependency.
	
	Therefore, if $c_1 = 0$ or $c_2 = 0$, $X_1$ and $X_2$ are conditionally independent given $Z=z$.
\end{IEEEproof}
}

\subsection{Proof of Theorem~\ref{thm:two_step_dependency}}
\label{sec:thm1_proof}

\begin{IEEEproof}
Let's firstly recall the following relationship between currents and voltages in a grid with $M$ buses. For bus $i$,
\begin{equation}
\label{eq:iv_p}	
\Delta I_i = \Delta V_iy_{ii} - \sum_{k \in \mathcal{N}(i)}\Delta V_ky_{ik},
\end{equation} 
with $y_{ii} = \sum_{k \in \mathcal{N}(i)}y_{ik} +\frac{1}{2}b_i$. Given $\Delta V_k = \Delta v_k$ for all $k \in \mathcal{N}(i) \cup \mathcal{N}^\RN{2}(i)$, the equation above becomes 
\begin{equation}
\label{eq:iv_p_rewrite}
\Delta I_i + \sum_{k \in \mathcal{N}(i)}\Delta v_ky_{ki} = \Delta V_iy_{ii}.
\end{equation}
This equation has only has two random variables, $\DV_i$ and $\DI_i$. We can rewrite the equation above as
\begin{equation}
	\label{eq:bus_i}
	\DI_i = a_i\DV_i + b_i,
\end{equation}
where $a_i$ and $b_i$ are constants.

For bus $l$ in $\mathcal{N}(i)$, given $\DV_k = \Dv_k$ for all $k \in \mathcal{N}(i) \cup \mathcal{N}^\RN{2}(i)$, we have a similar equation
\begin{equation}
	\label{eq:bus_l}
\Delta I_l + \sum_{k \in \mathcal{N}^\RN{2}(i)}\Delta v_ky_{ki} + \DV_iy_{il} = \Delta v_ly_{ll}.
\end{equation}
In (\ref{eq:bus_l}), the only unknown voltage variable is $\DV_i$ and therefore, $\DI_i$ and $\DI_l$ are conditionally dependent given $\Delta \mathbf{V}_{\mathcal{N}(i) \cup \mathcal{N}^\RN{2}(i)}$ for all $l \in \mathcal{N}(i)$.

For bus $p$ in $\mathcal{N}^\RN{2}(i)\backslash\mathcal{N}(i)$, given $\DV_k = \Dv_k$ for all $k \in \mathcal{N}(i) \cup \mathcal{N}^\RN{2}(i)$, we the following equation
\rev{
\begin{equation}
\Delta v_py_{pp} = \Delta I_p + \sum_{\substack{k \in \mathcal{N}(i) \cup \mathcal{N}^\RN{2}(i) \\ k \neq p}}\Delta v_ky_{kp} + \sum_{\substack{r \in \\\mathcal{M}\backslash\{\mathcal{N}(i) \cup \mathcal{N}^\RN{2}(i) \cup \{i\}\}}} \DV_ry_{rp}  \label{eq:bus_p}
\end{equation}
}
\rev{We can rewrite (\ref{eq:bus_p}) as 
\begin{equation}
	\label{eq:bus_p_rewrite}
	 \Delta I_p =
\sum_{\substack{r \in \\ \mathcal{M}\backslash\{\mathcal{N}(i) \cup \mathcal{N}^\RN{2}(i) \cup \{i\}\}}} \DV_r \widetilde{\mathbf{a}}_r +  \widetilde{b} 	
\end{equation}
If assuming the random vector $\mathbf{Y}$ in Lemma~\ref{lemma:linear_indept2} as $\mathbf{Y} = [\DV_i, \DV_{\mathcal{M}\backslash\{\mathcal{N}(i) \cup \mathcal{N}^\RN{2}(i) \cup \{i\}\}}]^T$, using (\ref{eq:iv_p_rewrite}) and (\ref{eq:bus_p_rewrite}), we can apply Lemma~\ref{lemma:linear_indept2} to show that $\DI_i$ and $\DI_p$ are conditionally independent given $\Delta \mathbf{V}_{\mathcal{N}(i) \cup \mathcal{N}^\RN{2}(i)}$ for all $p \in \mathcal{N}^\RN{2}(i)\backslash\mathcal{N}(i)$.} Similarly, for buses that are more than two steps away from bus $i$, we can observe that their incremental current injection $\DI_q$ and $\DI_i$ are conditionally independent, i.e., $\DI_i \perp \DI_q|\DV_{\mathcal{N}(i) \cup \mathcal{N}^\RN{2}(i)}$ for $q \in \mathcal{M}\backslash\{\mathcal{N}(i) \cup \mathcal{N}^\RN{2}(i)\}$.

\rev{
For any bus $q$ that is two more hops away from bus $i$, i.e., $q \in \mathcal{M}\backslash\{\mathcal{N}(i) \cup \mathcal{N}^\RN{2}(i) \cup \{i\}\}$, the nodal equation is 
\begin{equation}
	\label{eq:bus_q}
\DI_q = y_{qq}\DV_q + \sum_{k \in \mathcal{M}\backslash\{i,q,\mathcal{N}(i)\}} y_{qk} \DV_k.
\end{equation}
As demonstrated in the example of Theorem~\ref{thm:two_step_dependency}, the voltage at bus $q$ can be written as a summation of current injections and the voltage at bus $q$, i.e.,
\begin{equation}
	\label{eq:bus_q_i}
	\DI_q = c_q\DV_q + \sum_{k \in \mathcal{M}\backslash\{i,q,\mathcal{N}(i)\}} \mathbf{d}_k \DI_k,
\end{equation}
where $c_q$ and $\mathbf{d}$ are constants. As we have proved earlier, $\DI_i$ and $\DI_q$ are conditionally independent, given $\DV_{\mathcal{N}(i) \cup \mathcal{N}^\RN{2}{i}}$. Also, $\DI_i$ and $\DI_k$ are conditionally independent because $\DI_k$ are current injections of buses that are two or more hops away. Therefore, $\DI_i$ and $\DI_q + \sum_{k \in \mathcal{M}\backslash\{i,q,\mathcal{N}(i)\}}\DI_k$ are conditionally independent. Using (\ref{eq:bus_i}) and (\ref{eq:bus_q}), $\DV_i$ and $\DV_q$ are conditionally independent. This proof holds for every bus $q$ that is more than two hops away from bus $i$.
}
\end{IEEEproof}

\subsection{Proof of (\ref{eq:V_lasso})}
\label{sec:V_lasso}
In this section, we will show that with the increment of voltage magnitude $\Delta|V|$, the linear relationship expressed in (\ref{eq:v_linear}) still holds. Therefore, we can still use the lasso method to find the bus connectivity.

Since $y_{ik}$ and $\Delta V_i$ are all complex numbers, we can express them in polar form, i.e., $y_{ik} = |y_{ik}|\exp{j\phi_{ik}}$ and $\Delta V_i = \Delta |V_i|\exp{j\theta_i}$. Then, letting $\epsilon = \Delta I_i/y_{ii} $, (\ref{eq:v_linear}) becomes
\begin{eqnarray*}
	&& \Delta |V_i|\exp{j\theta_i} \\
	&=& \sum_{k \in \mathcal{N}(i)} \frac{|y_{ik}|}{|y_{ii}|}\exp{j(\phi_{ik}-\phi_{ii})}\Delta |V_k|\exp{j\theta_k} + \epsilon\\
	&=& \sum_{k \in \mathcal{N}(i)}\Delta |V_k|\frac{|y_{ik}|}{|y_{ii}|} \exp{j(\phi_{ik}-\phi_{ii}+\theta_k)} + \epsilon.
\end{eqnarray*}
Reorganizing the equation above, we have
\begin{eqnarray*}
	\Delta |V_i| &=& \sum_{k \in \mathcal{N}(i)}\Delta |V_k|\frac{|y_{ik}|}{|y_{ii}|} \exp{j(\phi_{ik}-\phi_{ii}+\theta_k-\theta_i)} \\
	&+&  \epsilon\exp{-j\theta_i} \\
	&=& \sum_{k \in \mathcal{N}(i)} \Delta |V_k| \gamma_{ik} + \tilde{\epsilon}.
\end{eqnarray*}
If bus $i$ and $k$ are not connected, $y_{ik} = |y_{ik}| = 0$. Therefore, $\gamma_{ik} = 0$ since the exponential term cannot yield zero. The lasso problem in (\ref{eq:V_lasso}) is an approximation of the equation above because we assume $\gamma_{ik}$ is a real number. However, this assumption does not affect the results because $|\gamma_{ik}|=0$ is the only solution for non-connected branch pairs.

\subsection{European Representative Distribution Networks}
\label{sec:eu_network}
In this section, we will briefly summarize the five representative distribution networks used in Section~\ref{sec:num}. For more details, please refer to \cite{pretticodistribution}. The topology maps below are duplicated from \cite{pretticodistribution} with several modifications.

\textit{LV\_urban} and \textit{LV\_suburban} systems represent the low voltage networks in urban and suburban areas respectively. \textit{LV\_suburban\_mesh} system in Fig.~\ref{fig:LV_semiurban_mesh} is a modified grid from \textit{LV\_suburban} system by adding additional branches to create loops. For all networks, the nodal voltage is 400V and all branches are underground. Bus 1 connects the LV grid to the substation with a 20kV/0.4kV transformer. \textit{LV\_large} network is an artificial distribution grid by combining 31 \textit{LV\_surban\_mesh} LV distribution grids. This network includes 3534 buses and 4030 branches. Bus 1 of 20 \textit{LV\_surban\_mesh} grids are connected at a common slack bus Bus 0. The rest 11 \textit{LV\_surban\_mesh} grids are connected at the end buses (e.g., Bus 50 and Bus 114) of the first 20 grids. 

\putFig{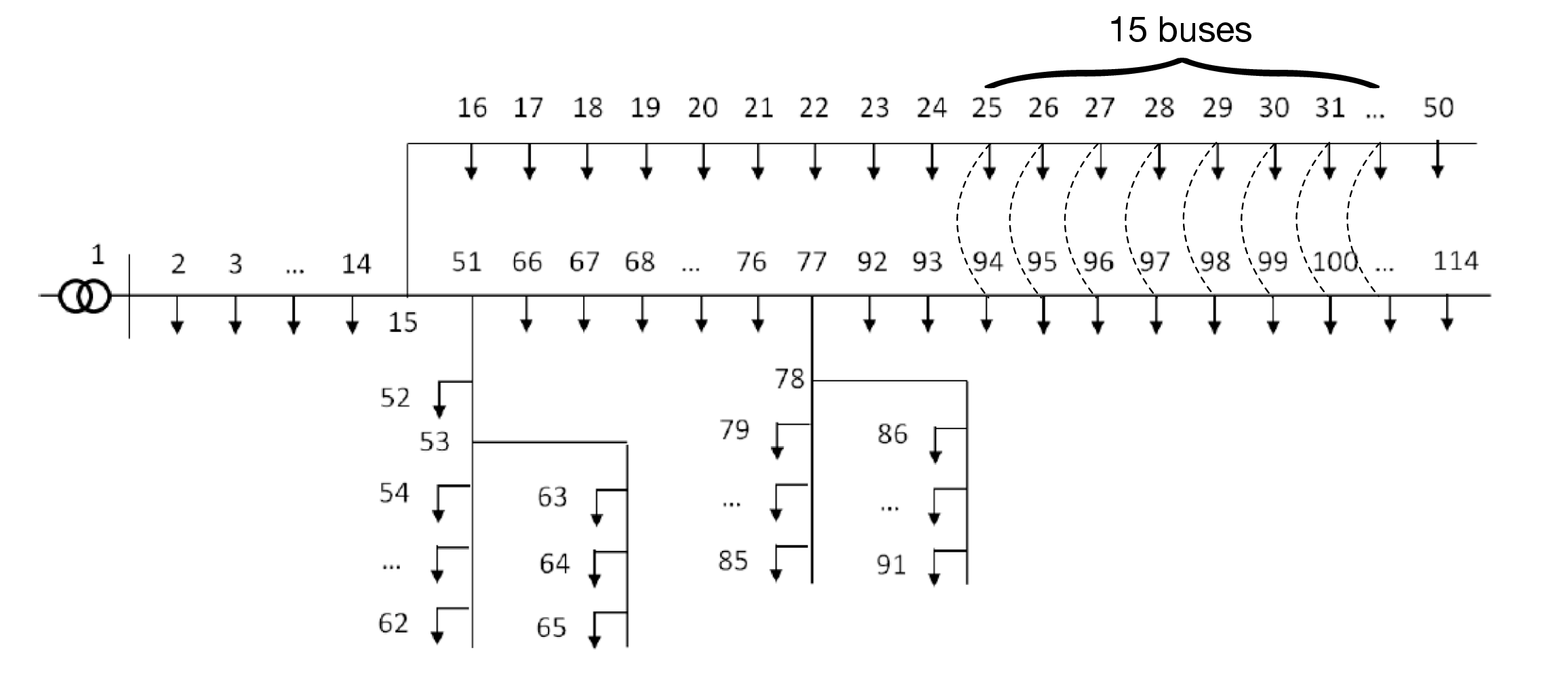}{Suburban Low Voltage Network (\textit{LV\_suburban}). The dashed lines indicate the branches that form suburban mesh low voltage network (\textit{LV\_suburban\_mesh}). All branches have the impedance $0.0019+j0.001\Omega$.}{\linewidth}

\textit{MV\_urban} and \textit{MV\_two\_substations} systems represent the medium voltage networks in urban and suburban areas. The bus voltage is 20kV and most branches are underground in these grids. Bus 1 connects with the HV/MV substation. Today, many MV distribution grids have mesh structures but radial operational topologies. Recently, several papers have shown that the closed-loop MV distribution grids can reduce power losses, provide a better voltage profile, and improve the power quality and service reliability \cite{chen2004feasibility,kim2013advanced}. Several utilities, such as Taipower, Florida Power Company, Hong Kong Electric Company, Singapore Power, and Korea Electric Power Cooperation, have operated mesh MV distribution grids in their service zone \cite{chen2004feasibility,pagel2000energizing,teo1995principles,jeon2016underground}. Also, as studied by \cite{celli2004meshed,de2014investigation}, for distribution grids with high DER penetration, MV distribution grids with mesh operational topology will be more reliable and efficient. In this study, our goal is to understand the performance of the proposed algorithm under both existing and future grid structures. Therefore, we validate our algorithm on MV grids with closed switches and closed-loop structure. \textit{MV\_urban\_mesh} system in Fig.~\ref{fig:MV_urban_mesh} are modified from \textit{MV\_urban} grid by adding additional branches. 

\putFig{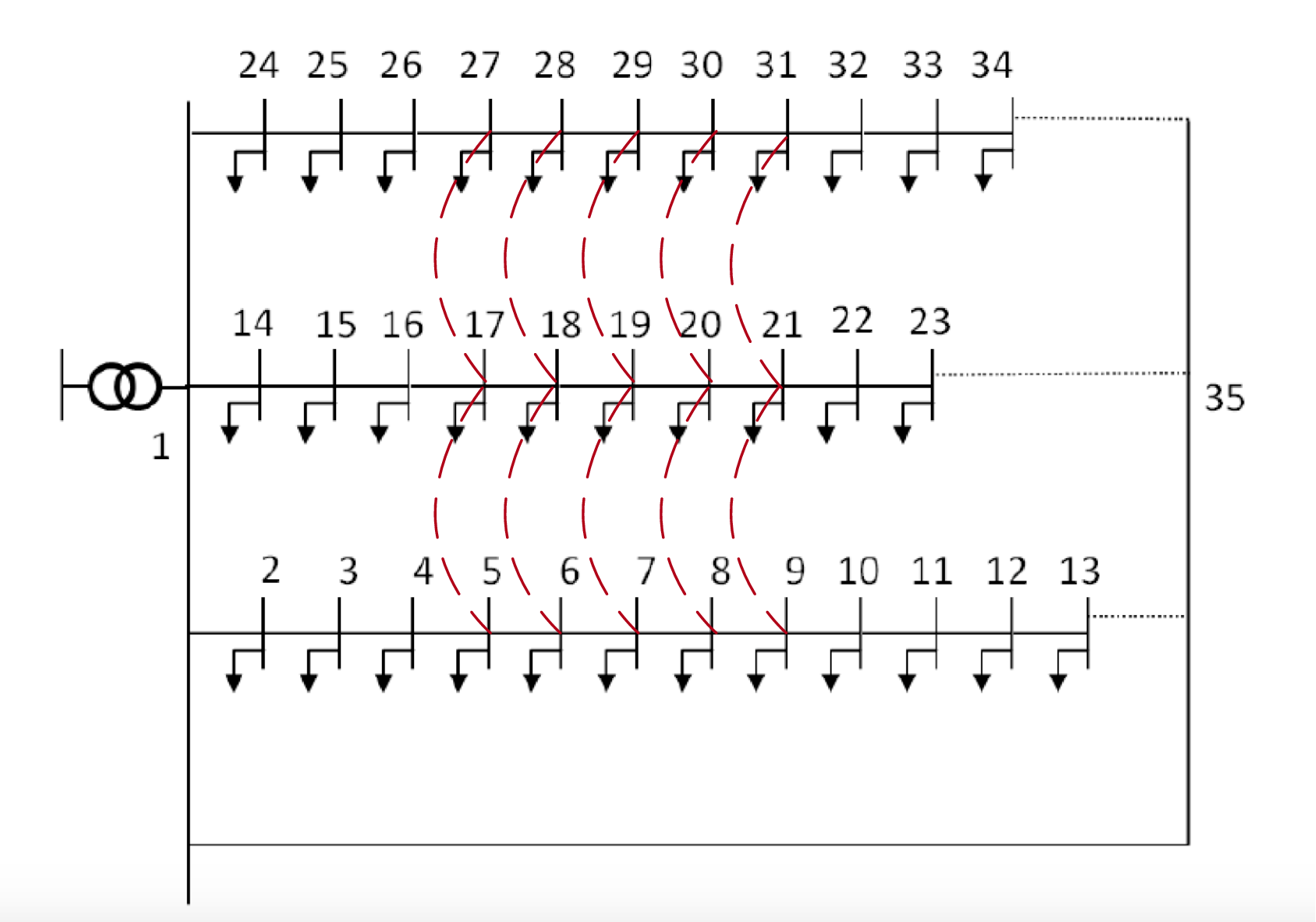}{Urban Medium Voltage Network (\textit{MV\_urban}). The red dashed lines indicate the branches and form urban mesh medium voltage network (\textit{MV\_urban\_mesh}). All branches have the impedance $0.0844 + j0.0444\Omega$.}{\linewidth}

\textit{MV\_rural} grid is a rural medium voltage network with four switches. Compared with urban MV networks, the distance of each branch in the rural network is much longer. To understand the impact of switch status, we close these switches and analyze the method performance on the loopy networks. 


\textit{Urban} system is an urban distribution grid, which includes MV feeders, LV feeders, MV/LV substations, and HV/MV substation. Most low voltage branches and all medium voltage branches are underground. To understand the impact of switch status, we close all switches and validate our method on the mesh networks.

Table~\ref{tab:rx} summarizes the $X/R$ ratios of all networks used in this paper. This table indicates that our algorithm works for typical medium and low voltage grids. Table~\ref{tab:avg_degree} shows the average and maximum number of neighbors per bus. These results show that the approach using $K$ top relative bus is valid.
\begin{table}[h!]
\centering
\caption{$X/R$ Ratio of the networks}
\label{tab:rx}
\begin{tabular}{|c||c|c|c|}
\hline
Grid                          & Minimum & Average & Maximum \\ \hline
123-bus                       & 0.49    & 1.66    & 2.36    \\ \hline
\textit{LV\_urban}            & 0.19    & 0.25    & 0.57    \\ \hline
\textit{LV\_suburban}         & 0.53    & 0.54    & 0.57    \\ \hline
\textit{MV\_urban}            & 0.52    & 0.52    & 0.53    \\ \hline
\textit{MV\_two\_substations} & 0.52    & 0.52    & 0.53    \\ \hline
\textit{MV\_rural}            & 0.92    & 0.93    & 0.93    \\ \hline
\textit{Urban}                & 0.07    & 0.39    & 3.48    \\ \hline
\end{tabular}
\end{table}

\begin{table}[H]
\caption{Average and Maximum Numbers of Neighbors}
\label{tab:avg_degree}
\centering
\begin{tabular}{|c||c|c|}
	\hline
	Grid & Average & Maximum  \\
	\hline
	123-bus & 1.98 & 4 \\
	\hline
	\textit{LV\_urban} & 1.85 & 4 \\
	\hline
	\textit{LV\_suburban} & 1.98 & 3 \\
	\hline
	\textit{MV\_urban} & 2.11 & 5 \\
	\hline
	\textit{MV\_two\_substations} & 2.04 & 3 \\
	\hline
	\textit{MV\_rural} & 2.03 & 9 \\
	\hline
	\textit{Urban} & 2.00 & 10 \\
	\hline
\end{tabular}
\end{table}

\bibliographystyle{IEEEtran}
\bibliography{ref}

\end{document}